\documentclass[11pt,a4paper]{article}
\usepackage[hyperref]{acl2021}

\usepackage{times}
\usepackage{graphicx}
\usepackage{wrapfig}
\usepackage{dsfont}
\usepackage{latexsym}
\usepackage[linesnumbered,algoruled,boxed,noend]{algorithm2e}
\usepackage{amssymb}
\usepackage{amsmath}
\usepackage{url} 
\SetKwInOut{Parameter}{parameter}
\usepackage{enumitem}
\usepackage{mathtools}
\usepackage{cleveref}
\usepackage{multirow}
\usepackage{hhline}
\usepackage[export]{adjustbox}
\usepackage{booktabs,array}
\usepackage{amsmath, bm}
\usepackage{color, colortbl}
\usepackage{xcolor}
\usepackage{nicematrix}

\usepackage[T2A,T1]{fontenc}
\usepackage[utf8]{inputenc}
\usepackage[russian,english]{babel}
\usepackage{textgreek}
\usepackage{CJKutf8}
\usepackage{dsfont}

\usepackage{resizegather}
\SetKwInput{KwInput}{Input}
\SetKwInput{KwOutput}{Output}
\newcolumntype{P}[1]{>{\centering\arraybackslash}p{#1}}
\usepackage{booktabs,makecell,tabularx} 
\setitemize{noitemsep,topsep=0pt,parsep=0pt,partopsep=0pt}
\let\oldnl\nl
\definecolor{Gray}{gray}{0.9}
\definecolor{LightCyan}{rgb}{0.88,1,1}
\newcommand{\nonl}{\renewcommand{\nl}{\let\nl\oldnl}}
\newcommand{\comm}[1]{{\nonl{\small{{\color{brown}{/*~#1}~*/}}}}}
\usepackage{pifont}
\newcommand{\corpusname}[0]{{\texttt{MickeyCorpus}}}
\newcommand{\taskname}[0]{{\textsc{MickeyProbe}}}%

\def\aclpaperid{1130}
\aclfinalcopy 
\usepackage{epigraph}

\usepackage{etoolbox}
\patchcmd{\epigraph}{\@epitext{#1}}{\itshape\@epitext{#1}}{}{}

\newlength{\Width}%
\newlength{\DepthReference}
\newlength{\HeightReference}
\newcommand{\MyColorBox}[2][red]%
{%
    \settowidth{\Width}{#2}%
    \colorbox{#1}%
    {%
        \raisebox{-\DepthReference}%
        {%
                \parbox[b][\HeightReference+\DepthReference][c]{\Width}{\centering#2}%
        }%
    }%
}
\begin{document}

\title{\vspace*{-0.5in}
{{\small \hfill ACL-IJCNLP 2021}\\
\vspace*{.25in}}
Common Sense Beyond English:
Evaluating and Improving \\ Multilingual Language Models for Commonsense Reasoning}

\author{
Bill Yuchen Lin \quad Seyeon Lee \quad Xiaoyang Qiao \quad Xiang Ren\\
\texttt{\{yuchen.lin, seyeonle,  xiaoyanq, xiangren\}@usc.edu}\\
Department of Computer Science and Information Sciences Institute,  \\ University of Southern California\\
}


\maketitle
\begin{abstract}
Commonsense reasoning research has so far been mainly limited to English.
We aim to evaluate and improve popular multilingual language models (ML-LMs) to help advance commonsense reasoning (CSR) beyond English.
We collect the Mickey corpus, consisting of 561k sentences in 11 different languages, which
can be used for \textit{analyzing} and \textit{improving} ML-LMs.
We propose Mickey Probe, a \textit{language-agnostic } probing task for fairly evaluating the common sense of popular ML-LMs across different languages.
Also, we create two new datasets, X-CSQA and X-CODAH, by translating their English versions to 15 other languages, so that we can evaluate popular ML-LMs for cross-lingual commonsense reasoning.
To improve the performance beyond English, 
we propose a simple yet effective method --- multilingual contrastive pretraining (MCP).
It significantly enhances sentence representations, yielding a large performance gain on both benchmarks (e.g., +2.7\% accuracy for X-CSQA over XLM-R$_L$)\footnote{We release our code and data at the project website: \url{https://inklab.usc.edu/XCSR/}.}.

\end{abstract}

\section{Introduction}
\label{sec:intro}
\begin{figure}[t]
    
	\centering 
	\includegraphics[width=1\linewidth]{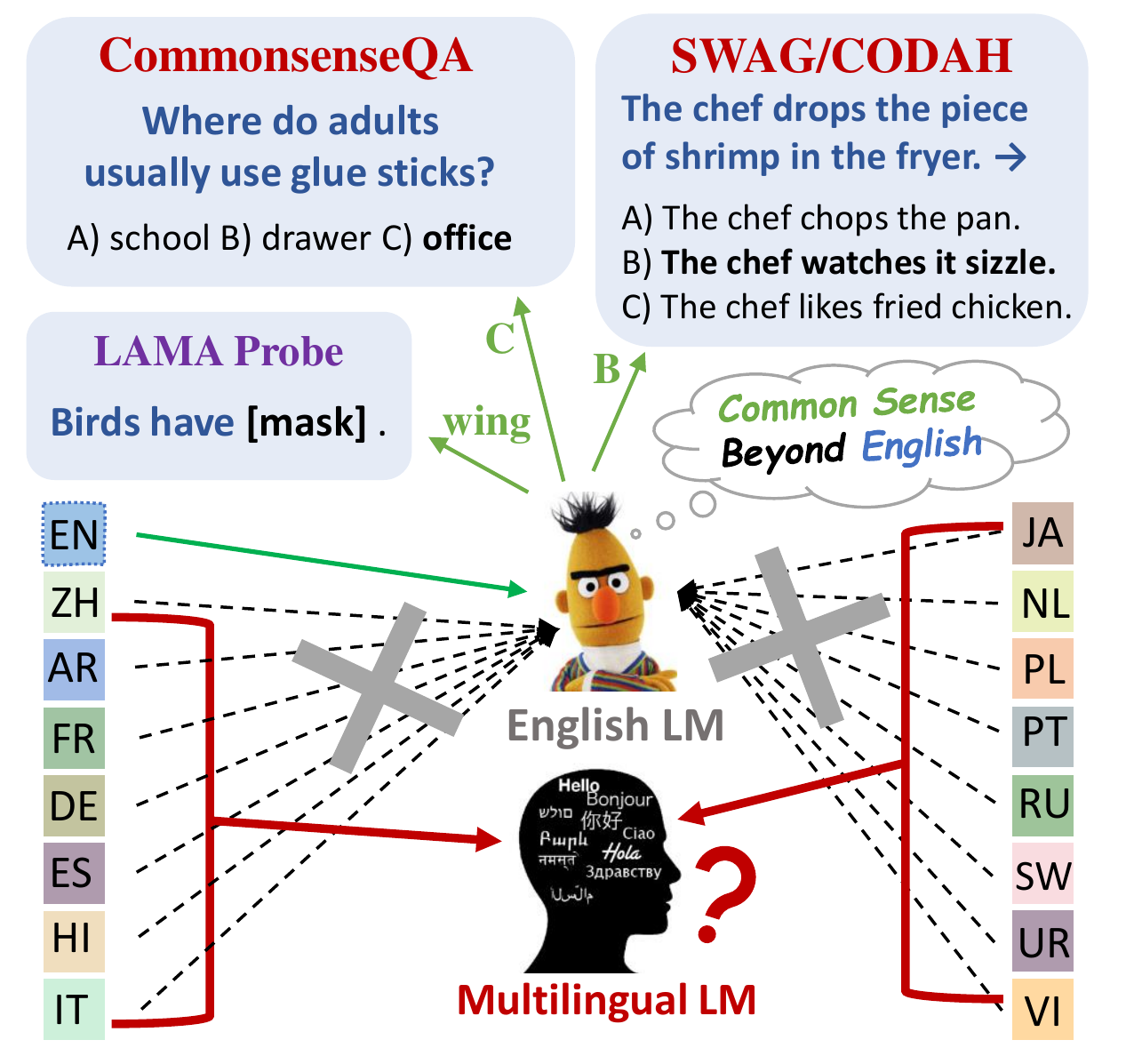}
	\caption{Commonsense reasoning is well-studied with benchmarks and LMs in English. Can we advance commonsense reasoning beyond English? }
	\label{fig:intro} 
\end{figure}

Understanding natural language relies heavily on commonsense reasoning (CSR), which is the process of making inferences with commonsense knowledge.
Commonsense knowledge is the set of general facts that reflect our natural understanding of the {physical world} 
and {human behavior},
which are usually seen as an implicit background when people communicate with each other using languages.
It is thus of vital importance to evaluate and improve the commonsense reasoning capability of language models (LMs), towards building general natural language understanding (NLU) systems~\cite{davis2015commonsense}.

Many recent benchmark datasets and probing methods have been proposed to evaluate machine common sense.
As shown in Figure~\ref{fig:intro}, the LAMA probe~\cite{petroni2019language} is for analyzing LMs' \textit{zero-shot} commonsense recalling ability; 
CommonsenseQA (CSQA)~\cite{talmor2018commonsenseqaaq} is instead a
multiple-choice QA task that needs fine-tuning; 
CODAH~\cite{Chen2019CODAHAA} and SWAG~\cite{zellers2018swagal} focus on the ability to complete the most plausible scenes.
However, all these works have been limited 
only to \textit{English}.
Consequently,
follow-up analysis and reasoning methods developed~\cite{kagnet-emnlp19, feng2020scalable, lin2020birds} also focus only on \textit{English} LMs like BERT~\cite{devlin2019}.
Such English-centric trend of commonsense reasoning studies not only limits our research scope,
but also tends to exacerbate English-specific bias that might prevent future methods from generalizing beyond English~\cite{ponti2020xcopa}.

It is of pressing urgency for the community to develop NLU systems that can serve \textit{all} languages in the world to bridge the gap between different cultures and eliminate language barriers~\cite{Hu2020},
and \textit{multilingual language models} (ML-LMs), such as XLM-R~\cite{conneau2019xlmr}, are among the most promising tools to achieve this ambitious goal.
Although ML-LMs have been evaluated in a few NLU tasks, e.g., XNLI~\cite{conneau2018xnli} and XTEMRE~\cite{Hu2020}, 
it is still relatively unclear how ML-LMs perform in commonsense reasoning tasks, due to the lack of 1) dedicated methods for probing common sense in ML-LMs and 2) multilingual benchmark datasets for commonsense reasoning.

To analyze how much common sense ML-LMs already have \textit{without any tuning},
we propose \taskname{}, a zero-shot probing task.
It tasks a ML-LM to rank a set of \textit{contrastive} assertions (i.e.,  declarative sentences) in the same language by their \textit{commonsense plausibility}, for which we use \textit{pseudo-likelihood} (PLL) ~\cite{salazar2020maskedlm} as a proxy.
Unlike the LAMA probe, 
it can study \textit{multi-token concepts} which are ubiquitous in some non-English languages.
In addition, it fairly compares performance across different languages via a \textit{language-invariant} evaluation protocol.
Alongside the probing task, we also create \corpusname{}, a large-scale multilingual dataset,
consisting of 561k sentences in 11 different languages.
Our experiments reveal that there are always large discrepancies across different languages in the tested ML-LMs, and different ML-LMs show very different language preferences.
Beyond supervision-free analysis of ML-LMs,
we also study their performance in commonsense reasoning tasks, such as CSQA and CODAH, within a \textit{cross-lingual transfer} setting (i.e., trained on English data and tested on other languages).
We find that existing ML-LMs tend to have much lower accuracy in commonsense reasoning beyond English.
We conjecture a major common weakness of existing ML-LMs is that their pretraining stages do not have a proper \textit{sentence-level} objective.
Therefore, we propose \textit{multilingual contrastive pre-training} (MCP), which tasks a ML-LM to select the correct assertion out of a set of $N$ contrastive assertions in $N$ \textit{different} languages.
We re-format \corpusname{} by sampling across languages and thus form a dedicated pre-training corpus for the MCP task.
To fairly evaluate different ML-LMs and validate the effectiveness of MCP, 
we create X-CSQA and X-CODAH, two cross-lingual commonsense reasoning datasets by translating their English versions to 15 other languages\footnote{The \textbf{16 languages} for X-CSQA and X-CODAH: \{en, zh, de, es, fr, it, jap, nl, pl, pt, ru, ar, vi, hi, \textit{sw}, \textit{ur}\}.}, including low-resource ones such as Swahili (\textit{sw}) and Urdu (\textit{ur}).
Experiments show that the proposed MCP objective indeed significantly improves the performance of state-of-the-art ML-LMs in cross-lingual commonsense reasoning.
Our contributions  are as follows:
\smallskip
\begin{itemize}
    \item \textbf{Resources.} We collect a large multilingual parallel corpus, \corpusname{}, consisting of 561k sentences in 11 languages, which can be used for \textit{analyzing} and \textit{improving} ML-LMs.
    We also create \texttt{X-CSQA} and \texttt{X-CODAH}, two cross-lingual CSR benchmarks in 16 languages, for question answering and scene completion, respectively.
     
    \item \textbf{Evaluation and analysis.}   We analyze multiple popular ML-LMs with \taskname{}, a \textit{language-invariant}, zero-shot task for probing common sense in ML-LMs; We also evaluate them on X-CSQA and X-CODAH in a cross-lingual transfer setting.
    
    \item \textbf{Method to improve ML-LMs.} We propose \textit{multilingual contrastive pretraining}, a simple and effective sentence-level pretext task for enhancing ML-LMs in cross-lingual commonsense reasoning, which significantly improves the state-of-the-art ML-LMs in cross-lingual commonsense reasoning.
\end{itemize}


 
\section{Background and Related Work}\label{sec:rel_work}
In this section, we introduce important concepts, background knowledge, and related work before we present our work in following sections.

\subsection{Multilingual Language Models}
\label{ssec:mllms}
A multilingual language model (ML-LM) aims to produce  text representations for multiple languages in a unified embedding space.
One of the unique advantages of ML-LMs is their potential ability to perform \textbf{zero-shot cross-lingual transfer} --- a model trained (or fine-tuned) on data in one language (usually English) can be directly used in other languages as well without further fine-tuning.
Improving ML-LMs is thus believed as one of the most promising approach towards multilingual NLU at scale. 
mBERT~\cite{devlin2019} is simply the BERT model~\cite{devlin2019} trained on multilingual corpora without specific designs about multilinguality.
The distil-mBERT (d-mBERT)~\cite{Sanh2019DistilBERTAD} is a smaller mBERT trained by knowledge distillation.
\citeauthor{lample2019xlm} (\citeyear{lample2019xlm}) proposed XLM(-100), which is pretrained with both masked language modeling (MLM) and translation language modeling (TLM). 
\citeauthor{conneau2019xlmr} (\citeyear{conneau2019xlmr}) further proposed XLM-R, which improves the XLM with a better sub-token vocabulary and high-quality multilingual corpora (CC100).
We leave the analysis of recent ML-LMs, such as mBART~\cite{mbart}, mT5~\cite{mt5}, and InfoXLM~\cite{chi-etal-2021-infoxlm} as future work.

Note that the above ML-LMs are pretrained only with
\textbf{token-level} training objectives such as MLM (i.e., recovering masked tokens in monolingual text) and TLM (i.e., recovering masked tokens in a pair of parallel sentences in two different languages).
However, most NLU tasks, including commonsense reasoning,  highly rely on \textbf{sentence-level} representations. 
We argue that a well-designed {sentence-level pre-training objective} should improve ML-LMs for NLU tasks.
This intuition motivates us to propose a sentence-level pre-training objective --- MCP (Section~\ref{sec:mcp}).

\subsection{Cross-lingual Language Understanding}
\label{ssec:xnlu}
There are a few recent multilingual benchmarks for NLU tasks, e.g., XTREME\cite{Hu2020}, TyDi QA\cite{clark2020tydi}, and XGLUE\cite{liang2020xglue}. 
XTREME and XGLUE  are unified large-scale multilingual multitask benchmarks, while Ty-Di QA focuses on the QA. 
These existing cross-lingual benchmarks have not covered \textit{commonsense reasoning tasks}, 
such as
CSQA~\cite{talmor2018commonsenseqaaq}, SWAG~\cite{zellers2018swagal}, and CODAH~\cite{Chen2019CODAHAA}.

CSQA is a question answering task and the other two are scene completion tasks, while all have a multiple-choice selection objective, as shown in Figure~\ref{fig:intro}.
These benchmarks are widely used to evaluate LMs for commonsense reasoning. 
Unfortunately, they are limited to English, not applicable to evaluate models of multilingual commonsense knowledge,
which motivates us to create X-CSQA and X-CODAH.
The goal of the recent {XCOPA}~\cite{ponti2020xcopa} dataset shares a similar goal, but it only focused on event-based causal reasoning in the scope of humans' social behavior, which is thus arguably more culturally biased.
In contrast, the X-CSQA and X-CODAH are mainly for evaluating general world knowledge and cover more fine-grained types of reasoning (e.g., quantitative, negation), and thus engage a more language-agnostic, comprehensive understanding of ML-LMs about common sense.



\subsection{The LAMA Probe and Its Limitations}
\label{ssec:LAMA}
The LAMA Probe~\cite{petroni2019language} is the seminal work on probing for common sense in (English) language models. It has a straightforward intuition: 
if a pretrained language model contains more commonsense knowledge, 
then it should be better at recalling a masked token in a commonsense assertion 
(e.g.,``\textit{birds have [mask]}'').
Specifically, given a LAMA-probe sentence $\boldsymbol{s}$ and its masked token ${w}_{t}$, 
a LM under testing uses all past and future tokens ---
$ \boldsymbol{s}_{\backslash t}:=\left({w}_{1}, \ldots, {w}_{t-1}, {w}_{t+1}, \ldots, {w}_{|\boldsymbol{s}|}\right)$.
as the input to rank all tokens in the vocabulary with the probability $P\left({w}_{t} \mid \boldsymbol{s}_{\backslash t}\right)$ via \textit{zero-shot} inference. 
One can evaluate the performance of recalling common sense by measuring the position of a correct token ``wing'' in the ranked list.
That is, the LAMA probe method uses \textbf{token-level probability} as a proxy to probe for common sense in LMs via ranking all tokens in their vocabularies.




This intuitive method, however, has several inherent limitations. 
First, in many other languages, \textit{multi-token concepts} are ubiquitous, for example, ``\begin{CJK*}{UTF8}{gbsn}图书馆\end{CJK*}'' (``library'' in Simplified Chinese).
\citeauthor{jiang2020x} (\citeyear{jiang2020x}) present several methods to decode multi-token entities so that they can adapt the LAMA probe to probe a LM for \textit{language-specific} analysis.
It is however infeasible to use token-level probing tasks if we want to analyze ML-LMs \textit{across languages}.
In addition, 
the evaluation metric of the LAMA probe could be unfair, because there can be many correct words for a masked position (e.g., ``\textit{birds have legs/eyes}'').
The ranking metrics of the LAMA probe, however, tend to ignore these facts, resulting in a less trustworthy analysis.
The vocabulary-specific ranking is unfair when comparing across different languages, as they can have very different label space.
These limitations of the LAMA Probe prevent us from analyzing common sense in ML-LM across topologically diverse languages.

We later found \citet{kassner-etal-2021-multilingual} proposed mLAMA, a
contemporaneous work extending the LAMA probes from English to other languages, via machine translation, sharing a similar goal to ours.
The mLAMA  focuses on factual knowledge about named entities via an entity-retrieval objective for mBERT only, while our \taskname{} aims to address commonsense knowledge via a  sentence-ranking objective for more ML-LMs.



\section{The Mickey Probe}
\label{sec:mickey}
The challenges of using the LAMA Probe for probing common sense in ML-LMs motivate us to propose a more suitable method for analyzing ML-LMs, one that can fairly compare across a diverse set of languages. 
We present \taskname, a \textit{M}ult\textit{i}lingual task for probing \textit{c}ommonsense \textit{k}nowledg\textit{e} and anal\textit{y}sis.
We design a language-agnostic  probing task with a sentence-selection objective for analyzing common sense of a ML-LM: 
given a set of assertions (i.e., declarative sentences) that have similar words and syntactic features, select the one with highest commonsense plausibility. 
We present the task formulation in this section and then introduce how we collect the dedicated dataset in Section~\ref{ssec:mickey_data}.

\paragraph{Notations.} 
We define a Mickey probe $M$ as a set of $K$ assertions in the same language, 
where one and only one of them (say, $M_i$) is the truth assertion with better commonsense plausibility than the other $K-1$ ones.
Each Mickey probe $M$ has multiple semantically equivalent versions in different languages.
Let us denote a \textit{language} by $l \in \mathcal{L}$ where $\mathcal{L}=\{en, fr, ru, zh, \dots\}$ and $|\mathcal{L}|$ is the number of languages of interest.
Then, 
$M^{l}$ is the probe $M$ in the language $l$.
For example, $M^{\text{en}}$ and  $M^{\text{fr}}$ denote the probes with the same meaning but in English (en) and French (fr) respectively.
We use $\mathcal{M}$ to denote a multilingual parallel dataset for \taskname, which consists of $T\times|\mathcal{L}|\times K$ assertions.
$T$ is the number of \taskname~items and each item has $K$ assertions and $|\mathcal{L}|$ language.
Finally, we can formally describe a multilingual parallel dataset $\mathcal{M}$ for \taskname:
\begin{align}
\begin{split}
\forall M\in \mathcal{M}, ~~&\forall (l_x,l_y)\in \mathcal{L}^2, ~~\forall i\in \mathbb{N}_{\leq K},  \\ 
    &{M}^{l_x}_i \bowtie {M}^{l_y}_i~.
\end{split}
\end{align}
We use the notation $\bowtie$ to indicate two assertions in different languages (e.g., $l_x$ and $l_y$) are semantically equivalent to each other. 
We leave the details of creating such an $\mathcal{M}$ in Section~\ref{ssec:mickey_data}.

\paragraph{Commonsense Probing Task. }
Given an instance ${M}$ for \taskname{} in the dataset $\mathcal{M}$, and suppose the index of the truth assertion to be $t$, 
a perfect multilingual language model would produce sentence probabilities such that it always gives the truth assertion ${M}^l_t$ the highest probability among other candidates for every language.
\begin{align}
\forall l\in \mathcal{L}, \forall i\in \mathbb{N}_{\leq K},~ P({M}^l_i) \leq P({M}^l_t).    
\end{align}

\begin{figure}[t]
	\centering 
	\includegraphics[width=1\linewidth]{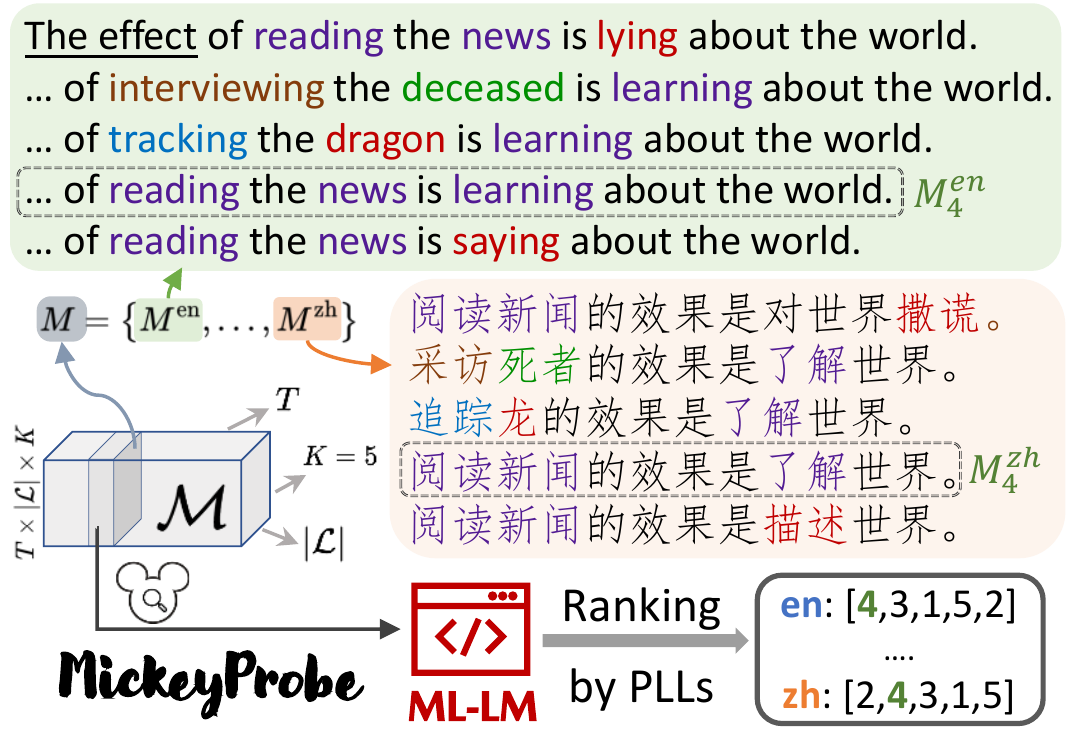}
	\caption{A Mickey Probe example $M$ has a set of probes in different languages (e.g., $M^{\text{en/zh}}$), and each of them is a set of 5 assertions. We rank assertions in the same language by their PLLs to probe common sense in ML-LMs across different languages. }
	\label{fig:mickeyprobe} 
\end{figure}

It is still an open problem to properly compute sentence probabilities from masked language models,
the recently proposed \textit{pseudo-log-likelihood scoring} (PLLs)~\cite{salazar2020maskedlm} has shown promising results in many downstream NLP applications that need sentence re-ranking (e.g., speech recognition, and translation), suggesting it is a promising proxy of sentence probability.
Given a sentence $\boldsymbol{s}$, 
its PLL is defined as:  
\begin{align}
\label{eq:pll}
      \log {P}(\boldsymbol{s}) = \operatorname{PLL}(\boldsymbol{s}) :=\sum_{i=1}^{|\boldsymbol{s}|} \log P \left({w}_{i} \mid \boldsymbol{s}_{\backslash i} \right)
\end{align}  
That is, we individually mask each token $w_i$ at a time and use the remaining context $\boldsymbol{s}_{\backslash i}$  to get the probability of a word $\boldsymbol{w_i}$ in the sentence $\boldsymbol{s}$.
Finally, we aggregate them to approximate $P(\boldsymbol{s})$.

\begin{table*}[th!] 
	\centering
	\scalebox{0.88
	}{
		\begin{tabular}{@{}c||c|c|c|c|c|c|c|c|c|c|c||c@{}}
\toprule
Models \textbackslash ~ $\mathcal{L}$ &
  \multicolumn{1}{c|}{\textbf{en}} &
  \multicolumn{1}{c|}{\textbf{de}} &
  \multicolumn{1}{c|}{\textbf{it}} &
  \multicolumn{1}{c|}{\textbf{es}} &
  \multicolumn{1}{c|}{\textbf{fr}} &
  \multicolumn{1}{c|}{\textbf{nl}} &
  \multicolumn{1}{c|}{\textbf{ru}} &
  \multicolumn{1}{c|}{\textbf{bg}} &
  \multicolumn{1}{c|}{\textbf{vi}} &
  \multicolumn{1}{c|}{\textbf{zh}} &
  \multicolumn{1}{c||}{\textbf{hi}} &
  \multicolumn{1}{c}{\textbf{avg}} \\ \midrule
BT-Cosine & 1.0 & 0.937 & 0.936 & 0.935 & 0.934 & 0.933 & 0.901 & 0.901 & 0.882 & 0.879 & 0.869 & 0.919 \\
CC-size (GB) & 300.8 & 66.6 & 30.2 & 53.3 & 56.8 & 29.3 & 278.0 & 57.5 & 137.3 & 46.9 & 20.2 & 97.9 \\ \midrule

\textit{Shortest} &  23.17\ & 27.21\ & 29.93\ & 31.00\ & 35.84\ & 31.68\ & 18.55\ & 22.01\ & 15.46\ & 25.07\ & 20.66\ & 25.51\ \\\midrule
d-mBERT & 62.95\ & 34.56\ & 25.26\ & 34.85\ & 50.46\ & 32.39\ & 21.49\ & 29.14\ & 19.77\ & 32.57\ & 25.88\ & 33.57\ \\
mBERT & 63.56\ & 35.58\ & 29.13\ & 44.70\ & 42.58\ & 35.15\ & 28.30\ & 36.03\ & 24.04\ & 28.15\ & 27.85\ & 35.92\ \\
XLM-100 & 60.57\ & 36.33\ & 26.49\ & 43.39\ & 32.53\ & 36.24\ & 32.90\ & 39.71\ & 25.79\ & 33.01\ & 31.49\ & 36.22\ \\
XLM-R$_B$ & 89.69\ & 58.94\ & 53.45\ & 60.88\ & 49.12\ & 59.99\ & 45.74\ & 45.26\ & 41.65\ & 51.02\ & 40.73\ & 54.22\ \\
XLM-R$_L$ & 90.03\ & 61.98\ & 53.42\ & 63.68\ & 59.47\ & 63.12\ & 50.03\ & 47.01\ & 45.30\ & 55.93\ & 43.98\ & 57.63\ \\\bottomrule

\end{tabular}
	} 
	
	\caption{The hit@1 accuracy (\%) of the five ML-LMs for the \taskname~ task. }
	\label{tab:results}
\end{table*}

\paragraph{Evaluation Metric.} 
The evaluation metric for \taskname~ over a multilingual parallel dataset $\mathcal{M}$ in a specific language $l$ is defined as the overall hit@k accuracy of the selection results
$\operatorname{hit@}{k}\operatorname{}(l) = {\sum_{M \in \mathcal{M}} \mathds{1} \{\operatorname{truth-rank}(M^l)  \leq k\} }~/~{|\mathcal{M}|}$
where $\operatorname{truth-rank}(M^l) $ means the the position of the truth assertion $M^l_t$ in $M^l$ sorted by their probabilities defined in Eq.~(\ref{eq:pll}).
The hit@1 is just equivalent to the conventional \textit{accuracy}.

\paragraph{Advantages of \taskname{}.} 
There are two key advantages of the \taskname{} for evaluating ML-LMs:
(1) The \textit{sentence-level probability} can be more generally applied in languages besides English, comparing with the LAMA probe which only studies single-token English words.
(2) The task formulation creates a relatively closed-ended setting, such that we can use a \textit{language-independent evaluation metric} to fairly compare across various languages within an ML-LM and compare across various ML-LMs for a particular language. 
In addition,
we can see LAMA Probe as a \textit{monolingual, word-level} version of the more general \taskname{}:
the LAMA Probe is when $\mathcal{L}=\{en\}$, and  $\{M^\text{en}\}=M \in \mathcal{M}$ is a \textit{huge} number of $K$ assertions (i.e., the vocabulary size) --- a \textit{fixed} \texttt{[mask]} is replaced by all tokens in the vocabulary.


\section{The Mickey Corpus and Evaluation}
\label{ssec:mickey_data}

We present a procedure for automatically creating a multilingual parallel dataset $\mathcal{M}$ for the probing task \taskname. 
Our collected corpus, named \corpusname~, has 561k sentences in 11 languages ($T=$10.2k, $K$=5, $|\mathcal{L}|$=11).

\subsection{Creating English Probes}
For the correct commonsense assertions in English, 
we have an existing resource, the OMCS corpus~\cite{Singh2002OpenMC} which contains human-written sentences in English that describe commonsense facts.
Each assertion can be used as a $M^{\text{en}}_t$ and we perform perturbations on it to create the other $K-1$ distractor assertions (i.e., false candidates), yielding an $M^{\text{en}}$ example.

Inspired by BERT-attack method~\cite{Li2020BERTATTACKAA}, 
we use a simple method to generate false assertions that are semantically related and syntactically similar to the truth assertions.
Given a correct assertion, we first randomly sample a few ($1\sim3$) words with a part-of-speech tag as noun, verb, or adjective, and replace them with [mask].
Then, we use a beam-search style method to decode the [mask] tokens one by one from left to right.
To ensure that the distractors are less plausible, 
we limit the decoding steps to only sample tokens that ranks between 200th$\sim$300th.
We repeat the above procedure multiple times with different sets of \texttt{[mask]} tokens.
Then, we use Stanza~\cite{qi2020stanza} to remove distractors that have sequences of POS tags or morphological features different from the truth assertions.
Finally, we sample $K-1$ of them as the distractors.

\begin{figure}[t]
	\centering 
	\includegraphics[width=1\linewidth]{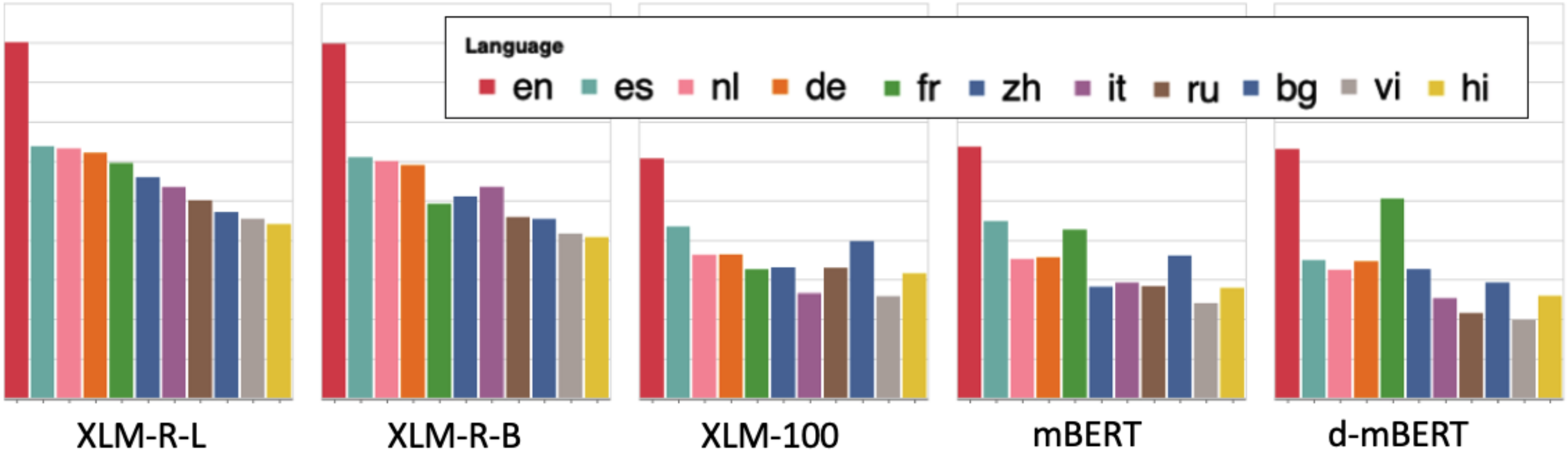}
	\caption{The \taskname~results in hit@1-acc. A larger version of this figure is in Appendix (Fig.~\ref{fig:mickeyresults_large}). }
	\label{fig:mickeyresults} 
\end{figure}

\subsection{Scaling to Ten Other Languages.}
We use \textit{bidirectional translation} with the MarianMT models~\cite{mariannmt} pretrained on the OPUS corpora~\cite{tiedemann2016opus}.
We translate all English probes to the 25 languages that has models in both directions and then translate them back to English.
As the outputs from these models might contain noise and errors,
we compute the semantic similarities (i.e., cosine similarity)
between the original $M^{\text{en}}$ and the back-translated $M^{\text{x-en}}$ via the  SentenceBERT~\cite{Reimers2019SentenceBERTSE} model.

To ensure the quality and fair comparisons,
we set a similarity threshold as 0.75 and keep the intersections of probes in all languages.
Considering some languages tend to have translations of lower quality, 
we finally choose the best 10 languages to build the Mickey Probe dataset for our analysis, yielding 10k examples in each language and 10.2k*5*11 $\approx$ 561k sentences in total.
The language set $\mathcal{L}=\{en, de, fr, ru, es, hi, vi, bg, zh, nl, it\}$.

Note that our purpose of checking the back-translation quality here is mainly to only keep the high-quality translations for all language pairs that we considered. 
Conventional metrics, e.g., BLUE score~\cite{Papineni2002BleuAM}, which focus on the \textit{exact} word match, are thus less suitable: given the original sentence ``I have a book'', the translation results ``I have a novel'' and ``I have a tool'' will be seen as equally wrong. 
Inspired by BERTScore~\cite{Zhang2020BERTScoreET}, the BT-cosine is based on SentenceBERT, which efficiently gives a higher score for the former and a lower score for the latter, due to the semantic relatedness between ``novel'' and ``book.''
We observed that most of our back-translations are in similar situations, and thus decide to use BT-cosine instead of others. 


%


\subsection{Analyzing ML-LMs with Mickey Probes}
\label{ssec:mickey_analysis}
We now use the \corpusname~
to evaluate the 5 pre-trained ML-LMs introduced in Section~\ref{ssec:mllms}: d-mBERT~\cite{Sanh2019DistilBERTAD}, mBERT~\cite{devlin2019}, XLM~\cite{lample2019xlm}, $\text{XLM-R}_\text{Base}$, and $\text{XLM-R}_\text{Large}$~\cite{conneau2019xlmr}.
All these ML-LMs pretraining objectives contain masked-word-prediction tasks, 
so we can easily use PPLs (Eq.~\ref{eq:pll}) to probe them {\textit{a zero-shot, supervision-free manner}} with hit@1 accuracy. ({The hit@2 results are shown in Appendix.})
We present a histogram in Figure~\ref{fig:mickeyresults} and show the concrete results in Table~\ref{tab:results}.
We find that there are always large discrepancies across different languages in all tested ML-LMs,
which motivates us to analyze the following questions.

\textit{\textbf{Q1: Do different ML-LMs have similar language preferences?}}
No.
We arrange the languages in all ML-LMs with the same order for Figure~\ref{fig:mickeyresults} --- the \textit{monotonically} descending order of XLM-R$_L$.
Interestingly, we find that different ML-LMs are good for different languages, 
resulting in a very diverse set of trends.
For example, XLM-R$_B$, has a higher performance in \textit{it} than \textit{zh} and \textit{fr},  unlike XLM-R$-L$ which are pre-trained on the same corpora with the same objectives.
mBERT and d-mBERT has stronger performance in \textit{fr} than \textit{nl} and \textit{de}, unlike XLM and XLM-R.

\textbf{\textit{Q2: Does length influence PLL ranking?}}
Not much. 
The PLL computation indeed tends to prefer shorter sequences (see Eq.~\ref{eq:pll}), so one may wonder if the length of assertions would influence the probing results.
The ``Shortest'' row in Table~\ref{tab:results} presents the results when we always select the shortest assertion within a probe, instead of  PLL ranking. 
The gaps between these scores and XLM-R-L's suggest that the probing task indeed uses PLL as a valid proxy for evaluating common sense based on sentence-level semantics.

\textbf{\textit{Q3: Is the translation quality a key factor?}}
We show ``BT-Cosine'', the mean of the cosine scores between the original English sentences and the back-translated ones,
and sort the table by these numbers.
The first 5 languages, \{de, it, es, fr, nl\} have the largest BT-Cosine, i.e., the best translation quality, 
and they indeed have better performances in general for XLM-R models. 
However, although \textit{zh} has a worse BT-score than \textit{vi}, all ML-LMs perform better in \textit{zh} than \textit{vi}.
Thus, we believe the translation quality of \corpusname{} will not be a factor to influence our understanding of ML-LMs.
Consequently, this suggests that further study must depend on pre-training corpora of each ML-LM in different languages.


\textbf{\textit{Q4: Does the size of pre-training corpora matter?}}
We list the size of the monolingual corpus in each language for CC-100 that XLM-R are pre-trained on (i.e., the CC-size row).
Although \textit{ru} has a much larger corpus than \textit{de}, \textit{it}, etc., 
the XLM-R performance in \textit{ru} is much worse.
In addition, \textit{fr} and \textit{nl} have almost the same translation quality while \textit{fr}'s CC-size is twice the size of \textit{nl}, but the performance in \textit{fr} is still much worse than \textit{nl}.
We conjecture this would be either due to the design of sub-token vocabulary or the text quality (instead of the size) of the CC-100 corpora. 

\paragraph{Further implications.} 
The benchmark results of five popular ML-LMs on the \taskname{} task over the \corpusname{} offer the initial and valuable understanding with a closer look at the commonsense knowledge of ML-LMs by probing them in a unified evaluation protocol.
One can either compare a ML-LM across different languages or compare a certain language across ML-LMs in Table~\ref{tab:results}. 
These comparable results support further analysis that can benefit the development of ML-LMs in the future.
After all, even the best ML-LM XLM-R$_L$ also degrades much in other languages, and also perform slightly worse than RoBERTa$_L$ in \textit{en} (93.4\%).
We argue (culture-invariant) common sense knowledge should be seen as an important way to connect multiple languages and thus better align them in a shared embedding space induced by a ML-LM.




\section{Multilingual Contrastive Pre-Training}
\label{sec:mcp}

In this section, 
we reformat the \taskname{}  so that we can reuse the \corpusname{}  for improving the pre-trained ML-LMs for commonsense reasoning beyond English.
We propose a {\textit{multilingual contrastive pre-training}} (MCP) task that focuses on enhancing the sentence-level representation of ML-LMs.
MCP improves a ML-LM in a \textit{multilingual}, \textit{contrastive} environment, where the model learns to select the assertion with the best commonsense plausibility from a set of contrastive sentences \textit{in different languages}.
Each MCP example is a set of \textit{multilingual} assertions while each Mickey probe is a \textit{monolingual} set.

\paragraph{MCP Dataset Creation from $\mathcal{M}$.}
We create pretraining examples for the MCP task by converting \taskname{} examples, as shown in the steps illustrated in Algorithm~\ref{alg:mcp}.
Simply put, we reformat a $K$-way Mickey Probe $M$ ($K\times|\mathcal{L}|$ assertions) to a MCP example by sampling a set of $V$  candidate assertions in $V$ different languages.
We convert all examples in the \corpusname{} $\mathcal{M}$ to build a new \textit{cross-lingual sentence-selection} dataset $\mathcal{C}$ for learning the MCP task.

\paragraph{MCP Learning.}
Given a MCP example $C\in\mathcal{C}$, 
we append one dense linear layer $f$ on top of a ML-LM with parameters denoted as $\Theta_{\text{ML-LM}}$ for learning to predict the \textit{commonsense plausibility score} of each assertion $C_i \in C$ as follows:

{  
  \begin{align}  
    \mathbf{h_i} &= \operatorname{ML-LM}(C_i).\texttt{[CLS]} \\
    o_i &= f(\mathbf{h_i}; \Theta_f) \\ 
    z_i&=\frac{e^{o_{i}}}{\sum_{j=1}^{V=|C|} e^{o_{j}}} \\
    \rho &= \sum_{i=1}^{V} - \mathds{1}_{i} \log \left(z_i\right)
\end{align}
}%

We first get the logit $o_i$ of each assertion by projecting its \texttt{[CLS]} embeddings $\mathbf{h_i}$ to a logit $o_i$ via a dense layer $f$ with parameters $\Theta_f$;
Then, we use SoftMax to normalize the logits as plausibility scores $z_i$;
Finally, we compute the cross-entropy loss $\rho$ where $\mathds{1}_i$=1 if $C_i$ is a correct assertion and $0$ otherwise.
We fine-tune $\{\Theta_{\text{ML-LM}}, \Theta_f\}$ to minimize the overall loss over the MCP dataset $\mathcal{C}$.

\begin{algorithm}[t]
	\nonl \textbf{In:}  {{$M\in \mathcal{M}$}} \comm{is a probe that has $|\mathcal L|$ sub-sets; each sub-set $M^{l_x}$ is a set of $K$ assertions in the same language $l_x\in \mathcal{L}$. $M^{l_x}_t$ is always the truth.}
	
	\nonl \textbf{Out}: $C$ \comm{A set of $V$ assertions in  different languages.}
	
	\nonl{\textbf{Remarks}: $\Gamma_{n}(X)$ is a function to randomly sample $n$ unique elements from a set $X$.} \\ ~\\
	\vspace{-1em}
	 $l_a \xleftarrow{}\Gamma_{1}(\mathcal{L})$ ~\comm{Pick an anchor language.} \\
	 $C \xleftarrow{} \{M^{l_a}_t\}$ \comm{Initiate w/ the truth assertion.} \\
	\comm{Iterate each sampled distractor language $l_i$.} \\ 
	\ForEach{$l_i \in \Gamma_{V-1}(\mathcal{L}-{l_a})$ } { 
	 \comm{Sample an index of distractor assertion.}\\
		$j\xleftarrow{}\Gamma_{1}(\mathbb{N}_{\leq K} - \{t\})$ \\
		\comm{Add a distractor assertion as a candidate.}\\
		${C}$\text{.add}($M^{l_i}_j$)
	}
	\caption{{Convert a Mickey Probe $M$ to an example for the MCP task.}}\label{alg:mcp}
\end{algorithm}



\section{Evaluation for Cross-lingual CSR}

In this section, we introduce the datasets, experimental setup, results, and our analysis.


\subsection{X-CSQA \& X-CODAH: Two New Benchmarks for Evaluating XCSR}

\begin{table*}[th!]
	\centering
	\scalebox{0.68
	}{
		\begin{tabular}{c|c|ccccccccc||cccccc||c}
	\toprule
 & \textbf{\underline{en}}  & \textbf{ de } & \textbf{ it } & \textbf{ es } & \textbf{ fr } & \textbf{ nl } & \textbf{ ru } & \textbf{vi } & \textbf{ zh } & \textbf{ hi } & \textbf{ pl } & \textbf{ ar } & \textbf{ ja } & \textbf{ pt } & \textbf{ \textit{sw} } & \textbf{\textit{ur} } & \textbf{ avg} \\ \midrule
   \rowcolor{gray!10} CC-size (GB) & 300.8 & 66.6 & 30.2 & 53.3 & 56.8 & 29.3 & 278.0 &  137.3 & 46.9 & 20.2 & 44.6 & 28.0 & 69.3 & 49.1 & 1.6 & 5.7 & 76.10 \\ 
 \midrule
 \rowcolor{cyan!30} \multicolumn{18}{c}{\textbf{X-CODAH} [\textit{Task:} Scene Completion;\quad \textit{Random Guess: }25.0;\quad \textit{RoBERTa$_L$ for \underline{en}: 81.6 }]}  \\  \midrule
mBERT & 42.9 & 33.1 & 33.5 & 33.8 & 35.2 & 33.7 & 31.9 & 22.8 & 38.0 & 26.5 & 31.0 & 34.8 & 34.0 & 37.2 & 30.8 & 31.5 & 33.2 \\
XLM-100 & 42.7 & 31.5 & 32.2 & 30.7 & 34.9 & 32.6 & 30.9 & 24.7 & 31.4 & 26.8 & 27.0 & 30.0 & 27.4 & 33.2 & 25.3 & 24.9 & 30.4 \\
XLM-R-B & \textit{50.1} & \textit{45.8} & \textit{44.4} & \textit{44.2} & \textit{45.2} & \textit{42.0} & \textit{44.1} & \textit{43.2} & \textit{44.6} & \textit{38.1} & \textit{41.9} & \textit{37.8} & \textit{42.0} & \textit{44.1} & \textit{35.6} & \textit{34.6} & \textit{42.4} \\
XLM-R-L  & \textbf{66.4} & \textbf{59.6} & \textbf{59.9} & \textbf{60.9} & \textbf{60.1} & \textbf{59.3} & \textbf{56.3} & \textbf{57.4} & \textbf{57.3} & \textbf{49.1} & \textbf{57.5} & \textbf{51.2} & \textbf{53.8} & \textbf{58.2} & \textbf{42.2} & \textbf{46.6} & \textbf{56.0} \\ \midrule
\textbf{MCP}{(XLM-R$_B$)} & \textit{52.2} & \textit{47.6} & \textit{46.2} & \textit{44.4} & \textit{48.1} & \textit{44.8} & \textit{42.9} & \textit{43.2} & \textit{45.7} & \textit{37.8} & \textit{41.8} & \textit{41.8} & \textit{42.9} & \textit{44.7} & \textit{37.2} & \textit{36.4} & \textit{43.6} \\
\textbf{MCP}{(XLM-R$_L$)} & \textbf{69.9} & \textbf{60.7} & \textbf{61.9} & \textbf{60.7} & \textbf{61.4} & \textbf{60.7} & \textbf{58.6} & \textbf{62.3} & \textbf{61.9} & \textbf{53.7} & \textbf{59.0} & \textbf{54.1} & \textbf{54.7} & \textbf{60.8} & \textbf{44.6} & \textbf{48.0} & \textbf{58.3} \\ \midrule
\rowcolor{blue!10} $\Delta(\text{XLM-R}_L)$ & +3.5 & +1.1 & +2.0 & -0.2 & +1.3 & +1.4 & +2.3 & +4.9 & +4.6 & +4.6 & +1.5 & +2.9 & +0.9 & +2.6 & +2.4 & +1.4 & +2.3\\
\midrule\midrule

  \rowcolor{purple!30}   \multicolumn{18}{c}{\textbf{X-CSQA} [\textit{Task:} Question Answering;\quad \textit{Random Guess: }20.0;\quad \textit{RoBERTa$_L$ for \underline{en}: }70.4 ]}     \\  \midrule 
mBERT & 38.8 & 29.6 & 36.4 & 35.3 & 33.8 & 32.6 & 32.7 & 22.2 & 37.8 & 21.1 & 27.2 & 27.7 & 31.4 & 34.1 & 21.8 & 23.7 & 30.4 \\
XLM-100 & 34.3 & 26.7 & 28.5 & 29.3 & 28.3 & 27.2 & 29.9 & 21.1 & 28.6 & 22.1 & 26.6 & 26.3 & 25.1 & 30.9 & 20.1 & 21.7 & 26.7 \\
XLM-R$_{B}$ & \textit{51.5} & \textit{44.1} & \textit{42.1} & \textit{44.8} & \textit{44.0} & \textit{43.3} & \textit{39.5} & \textit{42.6} & \textit{40.6} & \textit{34.6} & \textit{40.2} & \textit{38.4} & \textit{37.5} & \textit{43.4} & \textit{29.6} & \textit{33.0} & \textit{40.6} \\
XLM-R$_{L}$ & \textbf{66.7} & \textbf{56.1} & \textbf{58.2} & \textbf{59.5} & \textbf{60.3} & \textbf{56.8} & \textbf{52.1} & \textbf{51.4} & \textbf{52.7} & \textbf{48.7} & \textbf{53.9} & \textbf{48.4} & \textbf{50.0} & \textbf{59.9} & \textbf{41.6} & \textbf{45.2} & \textbf{53.8} \\ \midrule
\textbf{MCP}{(XLM-R$_B$)} & \textit{52.1} & \textit{46.2} & \textit{45.6} & \textit{44.3} & \textit{44.7} & \textit{45.3} & \textit{42.8} & \textit{45.3} & \textit{44.3} &\textit{ 36.8} & \textit{41.4} & \textit{36.8} & \textit{37.5} & \textit{44.9} & \textit{28.1} & \textit{33.4} & \textit{41.9} \\  
\textbf{MCP}{(XLM-R$_L$)} & \textbf{69.5} & \textbf{59.3} & \textbf{60.3} & \textbf{61.4} & \textbf{60.0} & \textbf{61.1} & \textbf{57.5} & \textbf{55.7} & \textbf{56.7} & \textbf{51.3} & \textbf{56.1} & \textbf{52.3} & \textbf{50.2} & \textbf{60.7} & \textbf{43.3} & \textbf{48.8} & \textbf{56.5} \\  \midrule
\rowcolor{blue!10} $\Delta(\text{XLM-R}_L)$ & +2.8 & +3.3 & +2.2 & +1.9 & -0.4 & +4.3 & +5.4 & +4.3 & +4.0 & +2.6 & +2.1 & +3.9 & +0.2 & +0.8 & +1.7 & +3.6 & +2.7 \\
\bottomrule
\end{tabular}
	} 
	
	\caption{Benchmark results for different ML-LMs and MCP-enhanced models for X-CSQA and X-CODAH in a zero-shot cross-lingual setting. $\Delta$ is the improvement of MCP. \{\textit{pl,ar,ja,pt,sw,ur}\} are unseen in MCP. }
	\label{tab:xcsr}
\end{table*}

To evaluate ML-LMs for commonsense reasoning in a cross-lingual zero-shot transfer setting, 
we create two benchmark datasets, namely X-CSQA and X-CODAH.
Table~\ref{tab:stat} shows the statistics of the two datasets.
Specifically, 
we use online commercial services such as \textit{DeepL Pro Translate} to collect high-quality translations of the examples in CSQA and CODAH for \textbf{15 languages} other than English.
The size of CODAH is small (only 2.7k), so we use 7k SWAG validation examples as additional training data which share the same formulation.
We discuss the reduction of \textit{cultural differences} and quality control of automatic translations as well as other details in \textit{Ethical Considerations} (the paragraph for cultural bias reduction) and \textit{Appendix} (A). 
As our goal is to evaluate different ML-LMs (instead of different languages) in a unified evaluation protocol for cross-lingual commonsense reasoning, we argue that such automatically translated examples, although might contain noise, can serve as a starting benchmark for us to obtain meaningful analysis before more human-translated datasets will be available in the future.

\begin{table}[t]
\centering
\scalebox{0.9}{
\begin{tabular}{@{}c||c|c@{}}
\toprule
Stat. $\downarrow$ Dataset $\rightarrow$ & {\MyColorBox[purple!10]{\textbf{X-CSQA}}}
 & {\MyColorBox[cyan!10]{\textbf{X-CODAH}}}     \\\midrule
Task Format            & QA     & SceneComp.   \\
\# Languages           & 15 + \textit{en}     & 15 + \textit{en}        \\
\# Options per Example            & 5      & 4   \\
\midrule
\# Training (en)   & 8,888   & \textit{8,476}        \\
\# Dev per Lang.       & 1,000   & 300           \\
\# Test  per Lang.      & 1,074   & 1,000          \\ \midrule
\# Total Instances            &  80,550     &  60,000 \\
\bottomrule
\end{tabular}
}
\caption{{Statistics of the two X-CSR datasets.} }
\label{tab:stat}
\end{table}

\subsection{Setup}
\label{ssec:xcsr_models}
We focus on 4 popular ML-LMs that we introduced in Section~\ref{ssec:mllms}: mBERT, XLM-100, XLM-R$_B$ and XLM-R$_L$ as well as our proposed MCP method.
For both tasks, 
we concatenate each prompt (the question or first sentence) and each of its options individually in the form of ``[CLS] prompt [SEP] option$_i$ [SEP]''.
Then, we fine-tune ML-LMs over the English training dataset and test them on other languages.

\textit{Why zero-shot cross-lingual transfer?}
It is almost impossible to collect data in \textit{all} languages that an NLU system might be used for.
Therefore, prior works mainly focus on zero-shot cross-lingual transfer~\cite{conneau2018xnli}, which is more meaningful and can offer \textit{lower-bound} performance analysis.
It is also an ideal setting for studying CSR because most commonsense facts are \textit{language-invariant}.
Thus, an English-finetuned ML-LM for CSR should be able to transfer its ability to a wide range of other languages as well.
Furthermore,
our goal of this paper is to evaluate and improve ML-LMs, so translating back to English and then use an English-only LM is also not helpful towards to this end.


\begin{figure}[t]
	\centering 
	\includegraphics[width=1\linewidth]{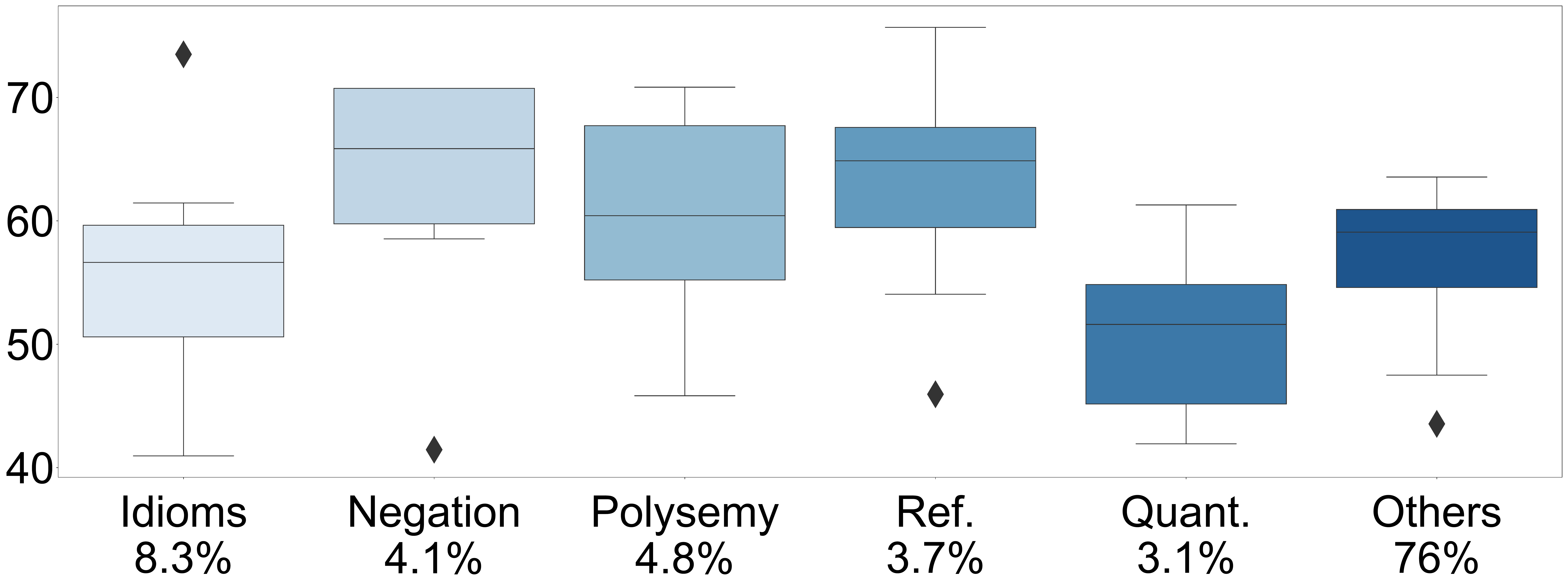}
	\caption{Categorized accuracy in for MCP(XLM-R$_L$) on X-CODAH. Each box is for 15 languages.  }
	\label{fig:boxplot} 
\end{figure}

\subsection{Experiments for Cross-lingual CSR}
\label{ssec:xcsr_exp}
In Table~\ref{tab:xcsr}, we present the empirical results over X-CODAH and X-CSQA for the ML-LMs as well as two models enhanced by our proposed MCP method. 
On both tasks, the XLM-R$_L$ performs the best with a large margin.
Enhanced by the MCP method, both XLM-R$_B$ and XLM-R$_L$ see significant improvement (e.g., 2.7\% absolute improvement for XLM-R$_L$ on X-CSQA-avg).

{\textit{\textbf{Can MCP's improvement generalize to unseen, low-resource languages?}}}
Note that MCP dataset only involves 9 languages here, and there are 6 languages that are totally \textit{unseen} in the MCP training (i.e., \{\textit{pl, ar, ja, pt, sw, ur}\}).
The largest performance gain is in \textit{ru} on X-CSQA and \textit{vi} on X-CODAH.
Surprisingly, we find the improvements on them are also large for XLM-R$_L$ (e.g., 48.4$\rightarrow$ 52.3 for \textit{ar}).
In addition, for the two \textit{low-resource} languages \textit{sw} and \textit{ur}, MCP also brings $2\sim3$ percentage points of improvement for XLM-R$_L$.
It is, however, not always the case for \textit{XLM-R}$_B$, which we conjecture tends to be more likely to overfit. 

\begin{figure}[t] 
	\centering 
	\includegraphics[width=1\linewidth]{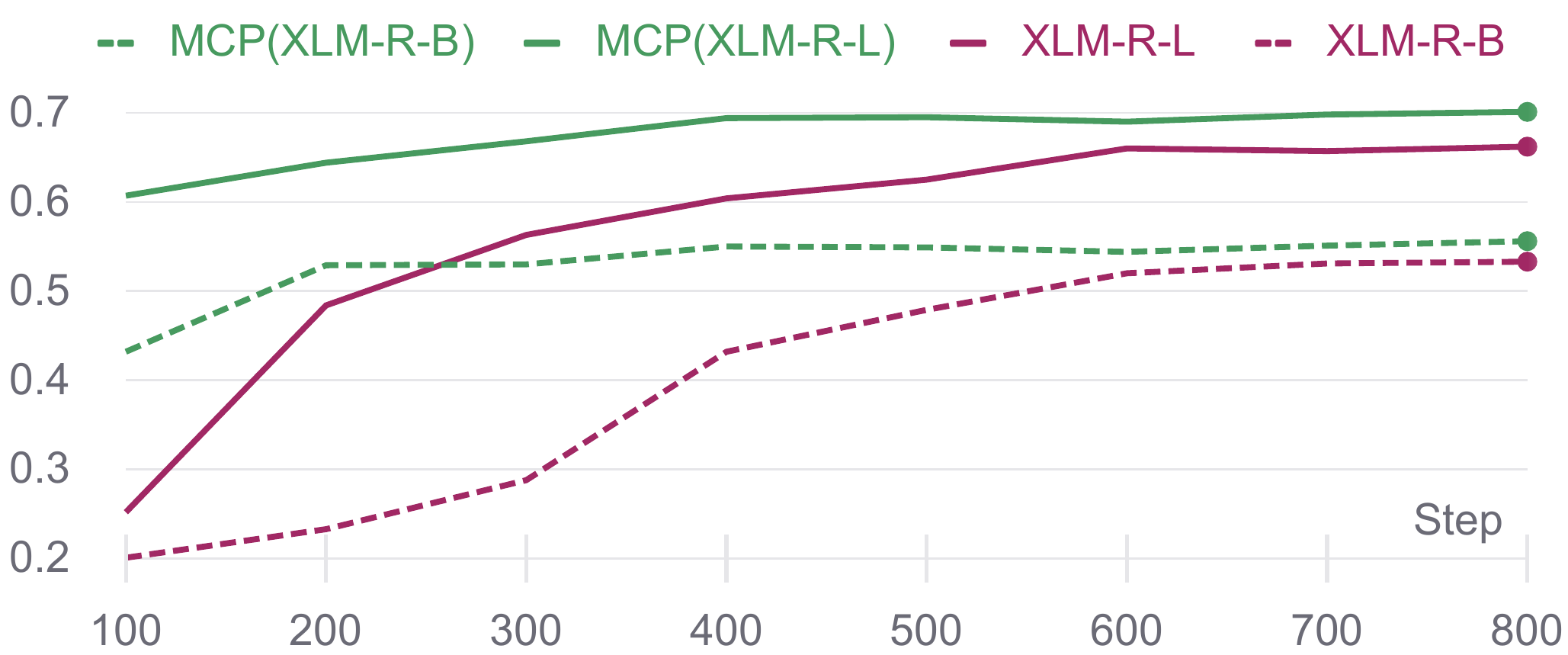}
	\caption{Dev acc v.s. learning steps on X-CSQA.  }
	\label{fig:devacc} 
\end{figure}

Although ML-LMs enjoy the merits of zero-shot cross-lingual transfer, their performances are usually \textit{worse} than the English-only RoBERTa$_L$ on the en-test (70.4\% vs 66.7\% for X-CSQA). 
Although MCP can mitigate the gap (70.4\% vs 69.5\%) for X-CSQA, there is still a large gap (81.6\% vs 69.9\%) for X-CODAH.
We use Fig.~\ref{fig:boxplot} to analyze how different categories of commonsense reasoning in CODAH~\cite{Chen2019CODAHAA} are diverse in different languages.
We find that \textit{others}, \textit{reference}, and \textit{negation} have relatively smaller variances across different languages, as they are more language-invariant.
However, a few \textit{polysemous}, \textit{idioms} examples can be English-specific which may not generalize to other languages. More detailed analysis is in Appendix.



From the curve of dev accuracy in Figure~\ref{fig:devacc}, we see that MCP-enhanced XLM-R models are much more \textit{sample efficient} and converge much faster than vanilla versions.
This suggests that the MCP, if used on a larger corpus with broader topics, can potentially produce a better ML-LM with more general usage, especially when only limited labelled is available.
Our results on XNLI-10\% (using 10\% of the training data)~\cite{conneau2018xnli} show that MCP-enhanced XLM-R$_L$ has 1.2 percent accuracy improvement on the average of 15 languages. 
As our focus in this paper is commonsense reasoning, we leave the study on other cross-lingual NLU tasks as future work. 
Importantly, our experiments imply that a proper (continual) pre-training task that has a (contrastive) sentence-level objective could improve both the final performance as well as learning efficiency.




\section{Conclusion}\label{sec:conclusion}
We evaluate and improve popular multilingual language models (ML-LMs) for advancing commonsense reasoning beyond English.
We propose the \taskname, a \textit{language-agnostic } probing task for analyzing common sense of ML-LMs in a zero-shot manner. 
With our proposed new benchmark datasets via automatic translation, X-CSQA and X-CODAH, we evaluate ML-LMs in a cross-lingual transfer setting for commonsense reasoning.
We also improve the state-of-the-art ML-LM with a simple yet effective method --- multilingual contrastive pre-training, which uses a sentence-level objective to enhance sentence representations, yielding a significant performance gain.
All above work is based on \corpusname{}, which can be used as both a probing dataset and a pre-training corpus for analyzing and improving ML-LMs.
We hope our resources and pre-training method for ML-LMs can help the community advance commonsense reasoning beyond English.

\section*{Acknowledgements}
This research is supported in part by the Office of the Director of National Intelligence (ODNI), Intelligence Advanced Research Projects Activity (IARPA), via Contract No. 2019-19051600007, the DARPA MCS program under Contract No. N660011924033 with the United States Office Of Naval Research, the Defense Advanced Research Projects Agency with award W911NF-19-20271, and NSF SMA 18-29268. The views and conclusions contained herein are those of the authors and should not be interpreted as necessarily representing the official policies, either expressed or implied, of ODNI, IARPA, or the U.S. Government. We would like to thank all the collaborators in USC INK research lab and the reviewers for their constructive feedback on the work.  

\section*{*~Ethical Considerations}

\paragraph{Resource Copyright} This work presents three new resources: \corpusname, X-CODAH, and X-CSQA, which are multilingual extension of the OMCS~\cite{Singh2002OpenMC}
\footnote{\url{https://github.com/commonsense/conceptnet5/wiki/Downloads}},
CSQA~\cite{talmor2018commonsenseqaaq}\footnote{\url{https://www.tau-nlp.org/commonsenseqa}}, and CODAH~\cite{Chen2019CODAHAA}\footnote{\url{https://github.com/Websail-NU/CODAH}} respectively.
All these three original sources of the data are publicly available for free, and we do not add any additional requirement for accessing our resources.
We will highlight the original sources of our data and ask users to cite the original papers when they use our extended versions for research.

\paragraph{Cultural Bias Reduction} Like most most multilingual parallel resources, especially in general NLU domain, there exists potential data bias due to the barrier of languages as well as \textit{cultural differences}~\cite{Acharya2020AnAO,lin2018miningcd}, which could induce the labeling differences on the same situation.
For example, a question like ``what do people usually drink in the morning? (coffee/tea/milk)'' or ``when does a wedding usually start? (morning/afternoon/evening)'' might be answered very differently by people from different backgrounds and cultures, not to mention different languages. 
The prior English commonsense resources which our datasets are built on are already possess such inherent bias, even with in the English language.
Therefore, before we translate CSQA and CODAH, we intentionally remove the examples that are either labeled as non-neutral by a pre-trained sentiment classifier, or contained any keywords that are relevant to social behavior (e.g., weddings).
We manually inspect test examples in X-CSQA and X-CODAH in the English and Chinese versions and have a strong confidence there is few strongly controversial example. 
However, we admit that such reduction of cultural differences in common sense has not been systematically measured in this work for other languages.


\subsection*{Application Risks of Cross-lingual CSR.} 
The work also evaluates a few multilingual language models (ML-LMs) for cross-lingual commonsense reasoning (XCSR), and introduced a new model which outperforms them.  
This raises the question of whether harm might arise from applications of XCSR---or more generally, since XCSR is intended as a step toward making English-only CSR more applicable in other languages, 
whether harm might arise more generally from existing ML-LMs.  
Among the risks that need to be considered in any deployment of NLP technology are that responses may be wrong or biased, in ways that would lead to improperly justified decisions.
Although in our view the current technology is still relatively immature, and unlikely to be fielded in applications that would cause harm of this sort, 
it is desirable that ML-LMs provide audit trails, 
and recourse so that their predictions can be explained to and critiqued by affected parties.  

\bibliography{citations_rebiber} 

\begin{thebibliography}{33}
\expandafter\ifx\csname natexlab\endcsname\relax\def\natexlab#1{#1}\fi

\bibitem[{Acharya et~al.(2020)Acharya, Talamadupula, and
  Finlayson}]{Acharya2020AnAO}
A.~Acharya, Kartik Talamadupula, and Mark~A. Finlayson. 2020.
\newblock An atlas of cultural commonsense for machine reasoning.
\newblock \emph{ArXiv}, abs/2009.05664.

\bibitem[{Chen et~al.(2019)Chen, D{'}Arcy, Liu, Fernandez, and
  Downey}]{Chen2019CODAHAA}
Michael Chen, Mike D{'}Arcy, Alisa Liu, Jared Fernandez, and Doug Downey. 2019.
\newblock \href {https://doi.org/10.18653/v1/W19-2008} {{CODAH}: An
  adversarially-authored question answering dataset for common sense}.
\newblock In \emph{Proceedings of the 3rd Workshop on Evaluating Vector Space
  Representations for {NLP}}, pages 63--69, Minneapolis, USA. Association for
  Computational Linguistics.

\bibitem[{Chi et~al.(2021)Chi, Dong, Wei, Yang, Singhal, Wang, Song, Mao,
  Huang, and Zhou}]{chi-etal-2021-infoxlm}
Zewen Chi, Li~Dong, Furu Wei, Nan Yang, Saksham Singhal, Wenhui Wang, Xia Song,
  Xian-Ling Mao, Heyan Huang, and Ming Zhou. 2021.
\newblock \href {https://www.aclweb.org/anthology/2021.naacl-main.280}
  {{I}nfo{XLM}: An information-theoretic framework for cross-lingual language
  model pre-training}.
\newblock In \emph{Proceedings of the 2021 Conference of the North American
  Chapter of the Association for Computational Linguistics: Human Language
  Technologies}, pages 3576--3588, Online. Association for Computational
  Linguistics.

\bibitem[{Clark et~al.(2020)Clark, Choi, Collins, Garrette, Kwiatkowski,
  Nikolaev, and Palomaki}]{clark2020tydi}
Jonathan~H. Clark, Eunsol Choi, Michael Collins, Dan Garrette, Tom Kwiatkowski,
  Vitaly Nikolaev, and Jennimaria Palomaki. 2020.
\newblock \href {https://doi.org/10.1162/tacl_a_00317} {{T}y{D}i {QA}: A
  benchmark for information-seeking question answering in typologically diverse
  languages}.
\newblock \emph{Transactions of the Association for Computational Linguistics},
  8:454--470.

\bibitem[{Conneau et~al.(2020)Conneau, Khandelwal, Goyal, Chaudhary, Wenzek,
  Guzm{\'a}n, Grave, Ott, Zettlemoyer, and Stoyanov}]{conneau2019xlmr}
Alexis Conneau, Kartikay Khandelwal, Naman Goyal, Vishrav Chaudhary, Guillaume
  Wenzek, Francisco Guzm{\'a}n, Edouard Grave, Myle Ott, Luke Zettlemoyer, and
  Veselin Stoyanov. 2020.
\newblock \href {https://doi.org/10.18653/v1/2020.acl-main.747} {Unsupervised
  cross-lingual representation learning at scale}.
\newblock In \emph{Proceedings of the 58th Annual Meeting of the Association
  for Computational Linguistics}, pages 8440--8451, Online. Association for
  Computational Linguistics.

\bibitem[{Conneau and Lample(2019)}]{lample2019xlm}
Alexis Conneau and Guillaume Lample. 2019.
\newblock \href
  {https://proceedings.neurips.cc/paper/2019/hash/c04c19c2c2474dbf5f7ac4372c5b9af1-Abstract.html}
  {Cross-lingual language model pretraining}.
\newblock In \emph{Advances in Neural Information Processing Systems 32: Annual
  Conference on Neural Information Processing Systems 2019, NeurIPS 2019,
  December 8-14, 2019, Vancouver, BC, Canada}, pages 7057--7067.

\bibitem[{Conneau et~al.(2018)Conneau, Rinott, Lample, Williams, Bowman,
  Schwenk, and Stoyanov}]{conneau2018xnli}
Alexis Conneau, Ruty Rinott, Guillaume Lample, Adina Williams, Samuel Bowman,
  Holger Schwenk, and Veselin Stoyanov. 2018.
\newblock \href {https://doi.org/10.18653/v1/D18-1269} {{XNLI}: Evaluating
  cross-lingual sentence representations}.
\newblock In \emph{Proceedings of the 2018 Conference on Empirical Methods in
  Natural Language Processing}, pages 2475--2485, Brussels, Belgium.
  Association for Computational Linguistics.

\bibitem[{Davis and Marcus(2015)}]{davis2015commonsense}
Ernest Davis and Gary Marcus. 2015.
\newblock Commonsense reasoning and commonsense knowledge in artificial
  intelligence.
\newblock \emph{Communications of the ACM}, 58(9):92--103.

\bibitem[{Devlin et~al.(2019)Devlin, Chang, Lee, and Toutanova}]{devlin2019}
Jacob Devlin, Ming-Wei Chang, Kenton Lee, and Kristina Toutanova. 2019.
\newblock \href {https://doi.org/10.18653/v1/N19-1423} {{BERT}: Pre-training of
  deep bidirectional transformers for language understanding}.
\newblock In \emph{Proceedings of the 2019 Conference of the North {A}merican
  Chapter of the Association for Computational Linguistics: Human Language
  Technologies, Volume 1 (Long and Short Papers)}, pages 4171--4186,
  Minneapolis, Minnesota. Association for Computational Linguistics.

\bibitem[{Feng et~al.(2020)Feng, Chen, Lin, Wang, Yan, and
  Ren}]{feng2020scalable}
Yanlin Feng, Xinyue Chen, Bill~Yuchen Lin, Peifeng Wang, Jun Yan, and Xiang
  Ren. 2020.
\newblock \href {https://doi.org/10.18653/v1/2020.emnlp-main.99} {Scalable
  multi-hop relational reasoning for knowledge-aware question answering}.
\newblock In \emph{Proceedings of the 2020 Conference on Empirical Methods in
  Natural Language Processing (EMNLP)}, pages 1295--1309, Online. Association
  for Computational Linguistics.

\bibitem[{Hu et~al.(2020)Hu, Ruder, Siddhant, Neubig, Firat, and
  Johnson}]{Hu2020}
Junjie Hu, Sebastian Ruder, Aditya Siddhant, Graham Neubig, Orhan Firat, and
  Melvin Johnson. 2020.
\newblock \href {https://sites.} {{XTREME: A Massively Multilingual Multi-task
  Benchmark for Evaluating Cross-lingual Generalization}}.
\newblock Technical report.

\bibitem[{Jiang et~al.(2020)Jiang, Anastasopoulos, Araki, Ding, and
  Neubig}]{jiang2020x}
Zhengbao Jiang, Antonios Anastasopoulos, Jun Araki, Haibo Ding, and Graham
  Neubig. 2020.
\newblock \href {https://doi.org/10.18653/v1/2020.emnlp-main.479} {{X}-{FACTR}:
  Multilingual factual knowledge retrieval from pretrained language models}.
\newblock In \emph{Proceedings of the 2020 Conference on Empirical Methods in
  Natural Language Processing (EMNLP)}, pages 5943--5959, Online. Association
  for Computational Linguistics.

\bibitem[{Junczys-Dowmunt et~al.(2018)Junczys-Dowmunt, Grundkiewicz, Dwojak,
  Hoang, Heafield, Neckermann, Seide, Germann, Aji, Bogoychev, Martins, and
  Birch}]{mariannmt}
Marcin Junczys-Dowmunt, Roman Grundkiewicz, Tomasz Dwojak, Hieu Hoang, Kenneth
  Heafield, Tom Neckermann, Frank Seide, Ulrich Germann, Alham~Fikri Aji,
  Nikolay Bogoychev, Andr{\'e} F.~T. Martins, and Alexandra Birch. 2018.
\newblock \href {https://doi.org/10.18653/v1/P18-4020} {{M}arian: Fast neural
  machine translation in {C}++}.
\newblock In \emph{Proceedings of {ACL} 2018, System Demonstrations}, pages
  116--121, Melbourne, Australia. Association for Computational Linguistics.

\bibitem[{Kassner et~al.(2021)Kassner, Dufter, and
  Sch{\"u}tze}]{kassner-etal-2021-multilingual}
Nora Kassner, Philipp Dufter, and Hinrich Sch{\"u}tze. 2021.
\newblock \href {https://www.aclweb.org/anthology/2021.eacl-main.284}
  {Multilingual {LAMA}: Investigating knowledge in multilingual pretrained
  language models}.
\newblock In \emph{Proceedings of the 16th Conference of the European Chapter
  of the Association for Computational Linguistics: Main Volume}, pages
  3250--3258, Online. Association for Computational Linguistics.

\bibitem[{Li et~al.(2020)Li, Ma, Guo, Xue, and Qiu}]{Li2020BERTATTACKAA}
Linyang Li, Ruotian Ma, Qipeng Guo, Xiangyang Xue, and Xipeng Qiu. 2020.
\newblock \href {https://doi.org/10.18653/v1/2020.emnlp-main.500}
  {{BERT}-{ATTACK}: Adversarial attack against {BERT} using {BERT}}.
\newblock In \emph{Proceedings of the 2020 Conference on Empirical Methods in
  Natural Language Processing (EMNLP)}, pages 6193--6202, Online. Association
  for Computational Linguistics.

\bibitem[{Liang et~al.(2020)Liang, Duan, Gong, Wu, Guo, Qi, Gong, Shou, Jiang,
  Cao, Fan, Zhang, Agrawal, Cui, Wei, Bharti, Qiao, Chen, Wu, Liu, Yang,
  Campos, Majumder, and Zhou}]{liang2020xglue}
Yaobo Liang, Nan Duan, Yeyun Gong, Ning Wu, Fenfei Guo, Weizhen Qi, Ming Gong,
  Linjun Shou, Daxin Jiang, Guihong Cao, Xiaodong Fan, Ruofei Zhang, Rahul
  Agrawal, Edward Cui, Sining Wei, Taroon Bharti, Ying Qiao, Jiun-Hung Chen,
  Winnie Wu, Shuguang Liu, Fan Yang, Daniel Campos, Rangan Majumder, and Ming
  Zhou. 2020.
\newblock \href {https://doi.org/10.18653/v1/2020.emnlp-main.484} {{XGLUE}: A
  new benchmark datasetfor cross-lingual pre-training, understanding and
  generation}.
\newblock In \emph{Proceedings of the 2020 Conference on Empirical Methods in
  Natural Language Processing (EMNLP)}, pages 6008--6018, Online. Association
  for Computational Linguistics.

\bibitem[{Lin et~al.(2019)Lin, Chen, Chen, and Ren}]{kagnet-emnlp19}
Bill~Yuchen Lin, Xinyue Chen, Jamin Chen, and Xiang Ren. 2019.
\newblock \href {https://doi.org/10.18653/v1/D19-1282} {{K}ag{N}et:
  Knowledge-aware graph networks for commonsense reasoning}.
\newblock In \emph{Proceedings of the 2019 Conference on Empirical Methods in
  Natural Language Processing and the 9th International Joint Conference on
  Natural Language Processing (EMNLP-IJCNLP)}, pages 2829--2839, Hong Kong,
  China. Association for Computational Linguistics.

\bibitem[{Lin et~al.(2020)Lin, Lee, Khanna, and Ren}]{lin2020birds}
Bill~Yuchen Lin, Seyeon Lee, Rahul Khanna, and Xiang Ren. 2020.
\newblock \href {https://doi.org/10.18653/v1/2020.emnlp-main.557} {{B}irds have
  four legs?! {N}umer{S}ense: {P}robing {N}umerical {C}ommonsense {K}nowledge
  of {P}re-{T}rained {L}anguage {M}odels}.
\newblock In \emph{Proceedings of the 2020 Conference on Empirical Methods in
  Natural Language Processing (EMNLP)}, pages 6862--6868, Online. Association
  for Computational Linguistics.

\bibitem[{Lin et~al.(2018)Lin, Xu, Zhu, and Hwang}]{lin2018miningcd}
Bill~Yuchen Lin, Frank~F. Xu, Kenny Zhu, and Seung-won Hwang. 2018.
\newblock \href {https://doi.org/10.18653/v1/P18-1066} {Mining cross-cultural
  differences and similarities in social media}.
\newblock In \emph{Proceedings of the 56th Annual Meeting of the Association
  for Computational Linguistics (Volume 1: Long Papers)}, pages 709--719,
  Melbourne, Australia. Association for Computational Linguistics.

\bibitem[{Liu et~al.(2020)Liu, Gu, Goyal, Li, Edunov, Ghazvininejad, Lewis, and
  Zettlemoyer}]{mbart}
Yinhan Liu, Jiatao Gu, Naman Goyal, Xian Li, Sergey Edunov, Marjan
  Ghazvininejad, Mike Lewis, and Luke Zettlemoyer. 2020.
\newblock \href {https://doi.org/10.1162/tacl_a_00343} {Multilingual denoising
  pre-training for neural machine translation}.
\newblock \emph{Transactions of the Association for Computational Linguistics},
  8:726--742.

\bibitem[{Papineni et~al.(2002)Papineni, Roukos, Ward, and
  Zhu}]{Papineni2002BleuAM}
Kishore Papineni, Salim Roukos, Todd Ward, and Wei-Jing Zhu. 2002.
\newblock \href {https://doi.org/10.3115/1073083.1073135} {{B}leu: a method for
  automatic evaluation of machine translation}.
\newblock In \emph{Proceedings of the 40th Annual Meeting of the Association
  for Computational Linguistics}, pages 311--318, Philadelphia, Pennsylvania,
  USA. Association for Computational Linguistics.

\bibitem[{Petroni et~al.(2019)Petroni, Rockt{\"a}schel, Riedel, Lewis, Bakhtin,
  Wu, and Miller}]{petroni2019language}
Fabio Petroni, Tim Rockt{\"a}schel, Sebastian Riedel, Patrick Lewis, Anton
  Bakhtin, Yuxiang Wu, and Alexander Miller. 2019.
\newblock \href {https://doi.org/10.18653/v1/D19-1250} {Language models as
  knowledge bases?}
\newblock In \emph{Proceedings of the 2019 Conference on Empirical Methods in
  Natural Language Processing and the 9th International Joint Conference on
  Natural Language Processing (EMNLP-IJCNLP)}, pages 2463--2473, Hong Kong,
  China. Association for Computational Linguistics.

\bibitem[{Ponti et~al.(2020)Ponti, Glava{\v{s}}, Majewska, Liu, Vuli{\'c}, and
  Korhonen}]{ponti2020xcopa}
Edoardo~Maria Ponti, Goran Glava{\v{s}}, Olga Majewska, Qianchu Liu, Ivan
  Vuli{\'c}, and Anna Korhonen. 2020.
\newblock \href {https://doi.org/10.18653/v1/2020.emnlp-main.185} {{XCOPA}: A
  multilingual dataset for causal commonsense reasoning}.
\newblock In \emph{Proceedings of the 2020 Conference on Empirical Methods in
  Natural Language Processing (EMNLP)}, pages 2362--2376, Online. Association
  for Computational Linguistics.

\bibitem[{Qi et~al.(2020)Qi, Zhang, Zhang, Bolton, and Manning}]{qi2020stanza}
Peng Qi, Yuhao Zhang, Yuhui Zhang, Jason Bolton, and Christopher~D. Manning.
  2020.
\newblock \href {https://doi.org/10.18653/v1/2020.acl-demos.14} {{S}tanza: A
  python natural language processing toolkit for many human languages}.
\newblock In \emph{Proceedings of the 58th Annual Meeting of the Association
  for Computational Linguistics: System Demonstrations}, pages 101--108,
  Online. Association for Computational Linguistics.

\bibitem[{Reimers and Gurevych(2019)}]{Reimers2019SentenceBERTSE}
Nils Reimers and Iryna Gurevych. 2019.
\newblock \href {https://doi.org/10.18653/v1/D19-1410} {Sentence-{BERT}:
  Sentence embeddings using {S}iamese {BERT}-networks}.
\newblock In \emph{Proceedings of the 2019 Conference on Empirical Methods in
  Natural Language Processing and the 9th International Joint Conference on
  Natural Language Processing (EMNLP-IJCNLP)}, pages 3982--3992, Hong Kong,
  China. Association for Computational Linguistics.

\bibitem[{Salazar et~al.(2020)Salazar, Liang, Nguyen, and
  Kirchhoff}]{salazar2020maskedlm}
Julian Salazar, Davis Liang, Toan~Q. Nguyen, and Katrin Kirchhoff. 2020.
\newblock \href {https://doi.org/10.18653/v1/2020.acl-main.240} {Masked
  language model scoring}.
\newblock In \emph{Proceedings of the 58th Annual Meeting of the Association
  for Computational Linguistics}, pages 2699--2712, Online. Association for
  Computational Linguistics.

\bibitem[{Sanh et~al.(2019)Sanh, Debut, Chaumond, and
  Wolf}]{Sanh2019DistilBERTAD}
Victor Sanh, Lysandre Debut, Julien Chaumond, and Thomas Wolf. 2019.
\newblock Distilbert, a distilled version of bert: smaller, faster, cheaper and
  lighter.
\newblock \emph{ArXiv}, abs/1910.01108.

\bibitem[{Singh et~al.(2002)Singh, Lin, Mueller, Lim, Perkins, and
  Zhu}]{Singh2002OpenMC}
Push Singh, Thomas Lin, Erik~T Mueller, Grace Lim, Travell Perkins, and Wan~Li
  Zhu. 2002.
\newblock Open mind common sense: Knowledge acquisition from the general
  public.
\newblock In \emph{OTM Confederated International Conferences" On the Move to
  Meaningful Internet Systems"}, pages 1223--1237. Springer.

\bibitem[{Talmor et~al.(2019)Talmor, Herzig, Lourie, and
  Berant}]{talmor2018commonsenseqaaq}
Alon Talmor, Jonathan Herzig, Nicholas Lourie, and Jonathan Berant. 2019.
\newblock \href {https://doi.org/10.18653/v1/N19-1421} {{C}ommonsense{QA}: A
  question answering challenge targeting commonsense knowledge}.
\newblock In \emph{Proceedings of the 2019 Conference of the North {A}merican
  Chapter of the Association for Computational Linguistics: Human Language
  Technologies, Volume 1 (Long and Short Papers)}, pages 4149--4158,
  Minneapolis, Minnesota. Association for Computational Linguistics.

\bibitem[{Tiedemann(2016)}]{tiedemann2016opus}
J{\"o}rg Tiedemann. 2016.
\newblock \href {https://www.aclweb.org/anthology/2016.eamt-2.8} {{OPUS} {--}
  parallel corpora for everyone}.
\newblock In \emph{Proceedings of the 19th Annual Conference of the European
  Association for Machine Translation: Projects/Products}, Riga, Latvia. Baltic
  Journal of Modern Computing.

\bibitem[{Xue et~al.(2021)Xue, Constant, Roberts, Kale, Al-Rfou, Siddhant,
  Barua, and Raffel}]{mt5}
Linting Xue, Noah Constant, Adam Roberts, Mihir Kale, Rami Al-Rfou, Aditya
  Siddhant, Aditya Barua, and Colin Raffel. 2021.
\newblock \href {https://www.aclweb.org/anthology/2021.naacl-main.41} {m{T}5: A
  massively multilingual pre-trained text-to-text transformer}.
\newblock In \emph{Proceedings of the 2021 Conference of the North American
  Chapter of the Association for Computational Linguistics: Human Language
  Technologies}, pages 483--498, Online. Association for Computational
  Linguistics.

\bibitem[{Zellers et~al.(2018)Zellers, Bisk, Schwartz, and
  Choi}]{zellers2018swagal}
Rowan Zellers, Yonatan Bisk, Roy Schwartz, and Yejin Choi. 2018.
\newblock \href {https://doi.org/10.18653/v1/D18-1009} {{SWAG}: A large-scale
  adversarial dataset for grounded commonsense inference}.
\newblock In \emph{Proceedings of the 2018 Conference on Empirical Methods in
  Natural Language Processing}, pages 93--104, Brussels, Belgium. Association
  for Computational Linguistics.

\bibitem[{Zhang et~al.(2020)Zhang, Kishore, Wu, Weinberger, and
  Artzi}]{Zhang2020BERTScoreET}
Tianyi Zhang, Varsha Kishore, Felix Wu, Kilian~Q. Weinberger, and Yoav Artzi.
  2020.
\newblock \href {https://openreview.net/forum?id=SkeHuCVFDr} {Bertscore:
  Evaluating text generation with {BERT}}.
\newblock In \emph{8th International Conference on Learning Representations,
  {ICLR} 2020, Addis Ababa, Ethiopia, April 26-30, 2020}. OpenReview.net.

\end{thebibliography}
\bibliographystyle{acl_natbib}

\clearpage

\appendix

\noindent
{\Large{\textbf{Appendix}}} \\
\smallskip
 

\section{Details for Dataset Construction}
Before we start the translation procedure, we first re-split the datasets of CSQA and CODAH such that the test set examples in the English language do not contain controversial examples or culture-related examples that would potentially cause cultural bias in our dataset. Please refer to the section of Ethical Considerations (following the Conclusion) in the main paper for more details.
Then, we use the DeepL Pro translation service to translate the 10 languages: \{de, fr, es, pt, it, nl, pl, ru, jap, zh\} and use Google Translation API to translate the others \{ar, sw, ur, vi, hi\}. 
\begin{figure*}[th!]
	\centering 
	\includegraphics[width=0.83\linewidth]{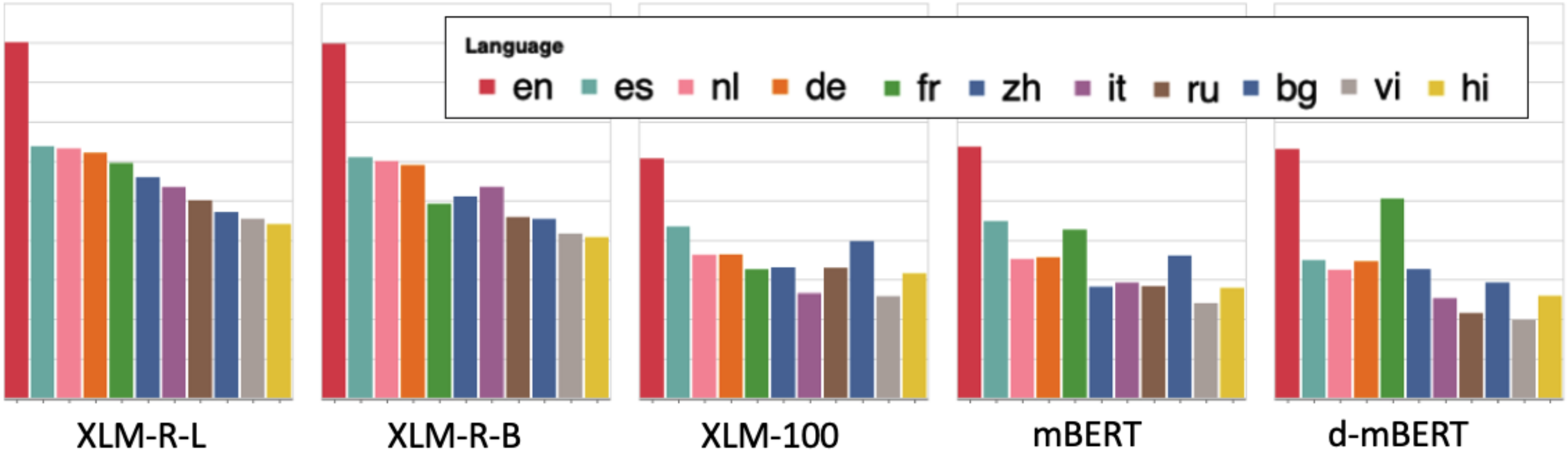}
	\caption{
	The \taskname~results in hit@1-acc. (An enlarged version of Figure~\ref{fig:mickeyresults}.)
	}
	\label{fig:mickeyresults_large} 
\end{figure*}

We agree that ideally we should use human experts to translate the examples in CSQA and CODAH, but the cost or building a large-scale multilingual dataset with the same scale of our datasets is extremely high -- around 10k USD.
As a matter of fact, most of the examples in CSQA and CODAH are very easy and short sentences, and most of them can be well translated by modern commercial translation APIs, because they usually have a hybrid system. 
Moreover, we choose the DeepL online service because it has been reported by many individual media as the best choice. 
To ensure the quality of the translation, 
we perform the translation for both directions and then use the same quality control method as we discussed in Section~\ref{ssec:mickey_data} for removing the examples that have lower cosine similarity between original English version and back-translated examples.
During the process, we manually went through the Chinese versions to find a suitable threshold for taking the intersection --- 0.85, which results in a comparable BT-cosine mean to the XNLI dataset~\footnote{We sampled 1k examples in the test set and follow the same procedure for the intersection language set. }.

\begin{table}[h!]
\centering
	\scalebox{0.68
	}{
		\begin{tabular}{@{}c||c|c|c|c|c|c|c|c@{}}
\toprule
Models  &
  \multicolumn{1}{c|}{\textbf{\#lgs}} &
  \multicolumn{1}{c|}{\textbf{tnz}} &
  \multicolumn{1}{c|}{\textbf{L}} &
  \multicolumn{1}{c|}{\textbf{H$_{m}$}} &
  \multicolumn{1}{c|}{\textbf{H$_{ff}$}} &
  \multicolumn{1}{c|}{\textbf{A}} &
  \multicolumn{1}{c|}{\textbf{V}} &
  \multicolumn{1}{c}{\textbf{\#para}}  \\ \midrule

mBERT       & 104   & WP    & 12 & 768  & 3072 & 12 & 110k & 172M     \ \\
XLM-100     & 100   & BPE          & 16 & 1280 & 5120 & 16 & 200k & 570M     \ \\
XLM-R$_{B}$ & 100   & SPM          & 12 & 768  & 3072 & 12 & 250k & 270M     \ \\
XLM-R$_{L}$       & 100   & SPM          & 24 & 1024 & 4096 & 16 & 250k & 550M \ \\\bottomrule

\end{tabular}
}
	\caption{Model Architectures. }
	\label{tab:modelarch}
\end{table}

\begin{table*}[h!]
	\centering
	\scalebox{0.85
	}{
		\begin{tabular}{c||c|c|c|c|c|c|c|c|c|c|c||c}
\toprule
Models \textbackslash ~ $\mathcal{L}$ &
  \multicolumn{1}{c|}{\textbf{en}} &
  \multicolumn{1}{c|}{\textbf{de}} &
  \multicolumn{1}{c|}{\textbf{it}} &
  \multicolumn{1}{c|}{\textbf{es}} &
  \multicolumn{1}{c|}{\textbf{fr}} &
  \multicolumn{1}{c|}{\textbf{nl}} &
  \multicolumn{1}{c|}{\textbf{ru}} &
  \multicolumn{1}{c|}{\textbf{bg}} &
  \multicolumn{1}{c|}{\textbf{vi}} &
  \multicolumn{1}{c|}{\textbf{zh}} &
  \multicolumn{1}{c||}{\textbf{hi}} &
  \multicolumn{1}{c}{\textbf{avg}} \\ \midrule

\rowcolor{gray!20} Shortest& 42.20 & 50.91 & 52.49 & 56.06 & 57.30 & 55.95 & 40.96 & 45.86 & 35.64 & 47.67 & 43.81 & 48.08\ \\\midrule
d-mBERT & 87.06 & 61.48 & 47.70 & 62.30 & 76.17 & 59.03 & 45.71 & 55.47 & 42.53 & 60.24 & 52.56 & 59.11\ \\
mBERT & 87.38 & 62.30 & 52.02 & 73.01 & 70.41 & 62.42 & 56.83 & 62.34 & 49.77 & 53.81 & 53.99 & 62.21\ \\
XLM-100 & 85.17 & 63.96 & 47.05 & 71.61 & 55.99 & 63.14 & 58.73 & 65.89 & 50.29 & 60.53 & 58.08 & 61.86\ \\
XLM-R$_B$ & 97.77 & 83.64 & 78.21 & 84.73 & 72.77 & 84.08 & 74.04 & 71.67 & 68.79 & 77.89 & 68.27 & 78.35\ \\
XLM-R$_L$ & 97.83 & 85.57 & 76.73 & 85.56 & 83.71 & 86.09 & 77.74 & 72.55 & 72.01 & 81.32 & 70.78 & 80.90\ \\\bottomrule

\end{tabular}
	} 
	
	\caption{The hit@2 accuracy of the five ML-LMs for the Mickey Probe task. }
	\label{tab:hit2acc}
\end{table*}

\begin{table*}[h!]
	\centering
	\scalebox{0.63
	}{
		\begin{tabular}{c||cc|ccccccccc||cccccc||c}
	\toprule
	\textbf{Category} &
 \textbf{RB} & \textbf{en}  & \textbf{ de } & \textbf{ it } & \textbf{ es } & \textbf{ fr } & \textbf{ nl } & \textbf{ ru } & \textbf{vi } & \textbf{ zh } & \textbf{ hi } & \textbf{ pl } & \textbf{ ar } & \textbf{ ja } & \textbf{ pt } & \textbf{ \textit{sw} } & \textbf{\textit{ur} } & \textbf{ avg} \\ 
 \midrule
Idioms & 79.52 & 69.88 & 61.45 & 56.63 & 60.24 & 73.49 & 60.24 & 57.83 & 50.6 & 55.42 & 45.78 & 59.04 & 50.6 & 50.6 & 56.63 & 44.58 & 40.96 & 55.87 \\
Neg. & 75.61 & 75.61 & 65.85 & 65.85 & 70.73 & 70.73 & 58.54 & 70.73 & 65.85 & 70.73 & 63.41 & 65.85 & 60.98 & 58.54 & 70.73 & 41.46 & 58.54 & 64.63 \\
Poly. & 79.17 & 75.00 & 58.33 & 66.67 & 68.75 & 70.83 & 60.42 & 66.67 & 68.75 & 56.25 & 54.17 & 60.42 & 45.83 & 66.67 & 68.75 & 45.83 & 50 & 61.46 \\
Ref.  & 86.49 & 78.38 & 62.16 & 67.57 & 67.57 & 64.86 & 64.86 & 67.57 & 62.16 & 54.05 & 67.57 & 72.97 & 75.68 & 45.95 & 54.05 & 62.16 & 56.76 & 64.02 \\
Quant. & 61.29 & 67.74 & 45.16 & 45.16 & 51.61 & 54.84 & 61.29 & 51.61 & 61.29 & 45.16 & 54.84 & 58.06 & 41.94 & 41.94 & 54.84 & 51.61 & 51.61 & 52.42
\\
Others & 82.89 & 68.95 & 61.05 & 62.37 & 59.74 & 59.08 & 60.66 & 57.37 & 63.03 & 63.55 & 53.29 & 57.89 & 54.08 & 55.13 & 60.79 & 43.55 & 47.5 & 58.00
\\
\midrule
\end{tabular}
	} 
	\caption{Benchmark results for MCP(XLM-R-L) on X-CODAH in different categories. RB = RoBERTa-Large.}
	\label{tab:xcsrco}
\end{table*}

\begin{table}[h!]
\centering
	\scalebox{0.8
	}{
\begin{tabular}{c||cccc}
\toprule
Model                             & \textbf{lr} & \textbf{\# epoch} & \textbf{\# wus} & \textbf{bsz} \\ \midrule
\multicolumn{5}{c}{X-CODAH}                                                                            \\ \midrule
\multicolumn{1}{c||}{mBERT}        & 3E-05      & 20              & 100            & 128             \\
\multicolumn{1}{c||}{XLM-100}      & 1E-05      & 20              & 100            & 64              \\
\multicolumn{1}{c||}{XLM-R-B}      & 1E-05      & 20              & 100            & 128             \\
\multicolumn{1}{c||}{XLM-R-L}      & 6E-06      & 10              & 100            & 64              \\ \midrule
\multicolumn{1}{c||}{MCP(XLM-R-B)} & 1E-05      & 20              & 100            & 128             \\
\multicolumn{1}{c||}{MCP(XLM-R-L)} & 6E-06      & 10              & 100            & 64              \\ \midrule
\multicolumn{5}{c}{X-CSQA}                                                                             \\ \midrule
\multicolumn{1}{c||}{mBERT}        & 3E-05      & 30              & 100             & 64              \\
\multicolumn{1}{c||}{XLM-100}      & 1E-05      & 20              & 300             & 64              \\
\multicolumn{1}{c||}{XLM-R-B}      & 1E-05      & 30              & 100             & 144             \\
\multicolumn{1}{c||}{XLM-R-L}      & 6E-06      & 10              & 100             & 64              \\ \midrule
\multicolumn{1}{c||}{MCP(XLM-R-B)} & 1E-05      & 30              & 100             & 144             \\
\multicolumn{1}{c||}{MCP(XLM-R-L)} & 6E-06      & 10              & 100             & 64              \\ \bottomrule
\end{tabular}
}
\caption{The optimal hyper-parameters for fine-tuning. (\textbf{lr} represents `learning rate'; training \textbf{\# epoch} ; \textbf{\# wus} = `\# warm up steps'; \textbf{bsz} = `batch size')}
	\label{tab:hyperparameters}
\end{table}

\section{Hyper-parameters}
We summarize hyper-parameters that we used for training ML-LMs on X-CODAH and X-CSQA in Table~\ref{tab:hyperparameters}. 
\textit{Evaluation Steps} are equally 100 for all models and datasets. \textit{Maximum Sequence Length} is 100 for X-CODAH and 64 for X-CSQA. The batch size here refers to ``\textit{train batch size per device} $\times$ \textit{\# GPUs} $\times$ \textit{\# gradient accumulation steps}''. 
Note that the MCP methods use the exactly the same hyper-parameters which we have found optimal by tuning over the dev set.
The learning rates we tried for all models are from the range \{3e-5, 2e-5, 1e-5, 8e-6, 6e-6, 5e-6\}. 
The warm up steps are selected from \{50, 100, 200, 300, 500\}.

\section{Details of ML-LMs }
Table~\ref{tab:modelarch} shows the model architectures and sizes that we used from \cite{conneau2019xlmr}. We show the tokenization (\textbf{tnz}) used by each Transformer model, the number of layers $L$, the number of hidden states of the model H$_{m}$, the dimension of the feed-forward layer H$_{ff}$, the number of attention heads $A$, the size of the vocabulary $V$ and the total number of parameters \#params.

\section{Additional Experimental Results}

\subsection{Hit@1 Accuracy in Histogram}

\subsection{Hit@k Accuracy of Mickey Probes}
Table~\ref{tab:hit2acc} shows the Hit@2 Accuracy of the five ML-LMs for the \textit{MickeyProbe}. 
Hit@2 Accuracy evaluates whether the models can rank the correct assertion within top 2. 
Unlike Hit@1 which only accepts best predictions, Hit@2 is more flexible. 
Thus, the performances in Hit@2 increase compared to the ones in Hit@1. We can see that the discrepancies across languages  still exist.






\subsection{Categorized X-CODAH Analysis}
Please refer the CODAH~\cite{Chen2019CODAHAA} paper for the definition and concrete examples in each category.
We show benchmark results of MCP(XLM-R$_L$) on X-CODAH within different carriages in Table~\ref{tab:xcsrco}. 
The \textbf{RB} stands for using the RoBERTa-Large model to fine-tune on the English X-CODAH dataset.
We find that the largest gaps in En are in the Idioms and the Others. Interestingly, we find that the quantities category is where MCP performs better than the RoBERTa large.


\end{document}


\title{Towards Common Sense Beyond English:\\ 
Evaluating and Improving Multilingual Language Models \\for Common Sense Reasoning }



\maketitle